\documentclass[runningheads]{llncs}

\usepackage{eccv}

\usepackage{eccvabbrv}

\usepackage{graphicx}
\usepackage{booktabs}

\usepackage[accsupp]{axessibility}  % Improves PDF readability for those with disabilities.

\usepackage{hyperref}

\usepackage{orcidlink}
\usepackage{makecell}
\usepackage{pifont}
\usepackage{bbding}
\definecolor{c1}{HTML}{0070C0}
\definecolor{c2}{HTML}{EC008C}
\usepackage{fontawesome}
\usepackage{wrapfig}
\usepackage{colortbl}
\usepackage{multirow} 

\begin{document}

% ---------------------------------------------------------------
% TODO REVIEW: Replace with your title
\title{M4-SAR: A Multi-Resolution, Multi-Polarization, Multi-Scene, Multi-Source Dataset and Benchmark for optical-SAR Object Detection} 

% TODO REVIEW: If the paper title is too long for the running head, you can set
% an abbreviated paper title here. If not, comment out.
\titlerunning{M4-SAR}

% TODO FINAL: Replace with your author list. 
% Include the authors' OCRID for the camera-ready version, if at all possible.
\author{Chao Wang\inst{1}\orcidlink{0000-0003-3766-8082} \and
Wei Lu\inst{2}\orcidlink{0009-0004-5197-5753} \and
Xiang Li\inst{3}\orcidlink{0000-0002-4996-7365} \and
Jian Yang\inst{1}\orcidlink{0000-0003-4800-832X} \and
Lei Luo\inst{1}\textsuperscript{\Envelope}\orcidlink{0000-0002-9976-0442}}
% \author{Chao Wang\inst{1} \and
% Wei Lu\inst{2} \and
% Xiang Li\inst{3} \and
% Jian Yang\inst{1} \and
% Lei Luo\inst{1}\textsuperscript{\Envelope}}

% TODO FINAL: Replace with an abbreviated list of authors.
\authorrunning{C. Wang et al.}
% First names are abbreviated in the running head.
% If there are more than two authors, 'et al.' is used.

% TODO FINAL: Replace with your institution list.
\institute{Nanjing University of Science and Technology, Nanjing 210094, China \and
Anhui University, Hefei 230601, China \and Nankai University, Tianjin 300350, China \\
Project page: \url{https://github.com/wchao0601/M4-SAR} \\
\email{wchao0601@163.com, luoleipitt@gmail.com}}
\maketitle
\let\thefootnote\relax\footnotetext{\textsuperscript{\Envelope} Corresponding Author}

\begin{abstract}
  Single-source remote sensing object detection using optical or SAR images struggles in complex environments. Optical images offer rich textural details but are often affected by low-light, cloud-obscured, or low-resolution conditions, reducing the detection performance. SAR images are robust to weather, but suffer from speckle noise and limited semantic expressiveness. Optical and SAR images provide complementary advantages, and fusing them can significantly improve the detection accuracy. However, progress in this field is hindered by the lack of large-scale, standardized datasets. To address these challenges, we propose a new comprehensive dataset for optical-SAR fusion object detection, named \textbf{M}ulti-resolution, \textbf{M}ulti-polarization, \textbf{M}ulti-scene, \textbf{M}ulti-source \textbf{SAR} dataset (\textbf{M4-SAR}). It contains 112,174 instance-level aligned image pairs and nearly one million labeled instances with arbitrary orientations, spanning six key categories. To enable standardized evaluation, we develop a unified benchmarking toolkit that integrates six state-of-the-art multi-source fusion methods. Additionally, we propose E2E-OSDet, a novel end-to-end multi-source fusion detection framework that mitigates cross-domain discrepancies and establishes a robust baseline for future studies. Extensive experiments on M4-SAR demonstrate that fusing optical and SAR data can improve mAP by 5.7\% over single-source inputs, with particularly significant gains in complex environments.
\end{abstract}

\section{Introduction}
\label{sec:intro}
\begin{figure*}[t]	
    \centering
    \includegraphics[width=0.98\linewidth]{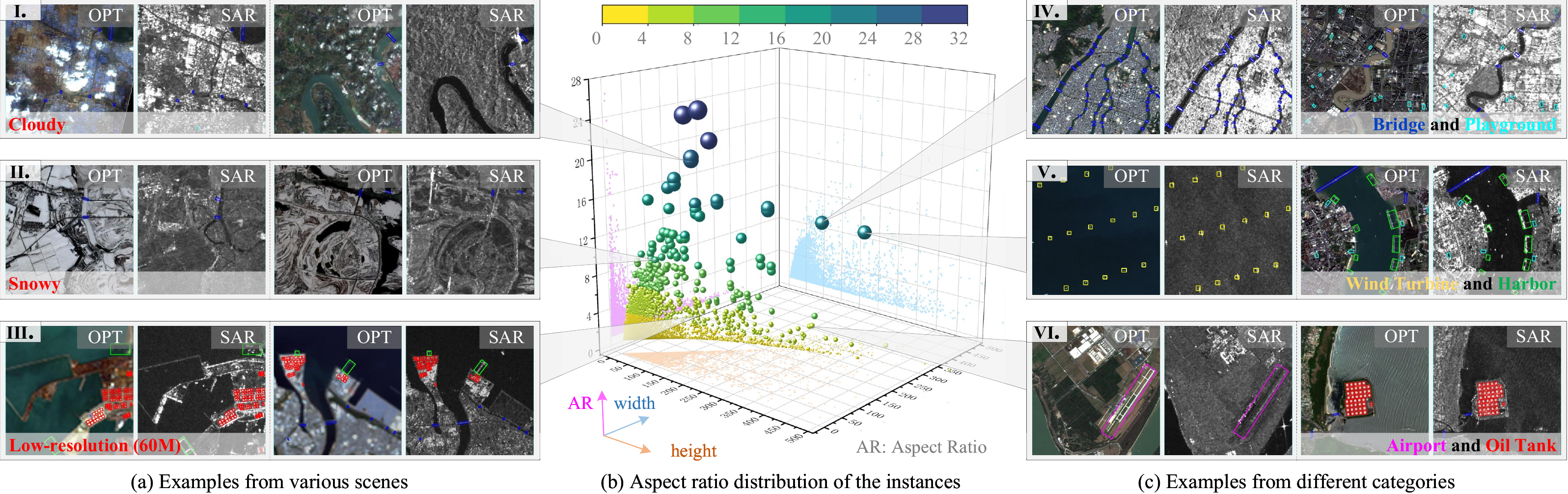}
    \caption {Examples of scenes and six key target categories in the proposed M4-SAR dataset, accompanied by instance size and aspect ratio distributions.
    }
    \label{motivation}
\end{figure*}

With the rapid development of deep learning techniques, single-source object detection (e.g., optical \cite{li2024lsknet,lu2025lwganet,lu2026unravelnet,girshick2015fast,he2017mask,li2024unleashing,Controllable-LPMoE,GLCONet} or synthetic aperture radar (SAR) \cite{wang2025msod,wang2026localized,MMRotate_Contributors_OpenMMLab_rotated_object_2022,geng2023focusing}), and multi-source object detection (optical-SAR \cite{li2021deep, zhang2022domain}) have garnered significant attention in the remote sensing field. These methods have demonstrated remarkable effectiveness in applications, including disaster monitoring and urban planning. As shown in Fig. \ref{motivation} (c), optical images provide high-resolution visual cues with rich texture, color, and spectral information, facilitating fine-grained feature extraction and semantic interpretation under clear atmospheric conditions. However, their detection performance degrades markedly under low-light, cloud-obscured, or low-resolution conditions (see Fig. \ref{motivation} (a)). Despite these drawbacks, the intuitive appearance and broad availability of optical data have fueled significant advancements in optical image object detection \cite{liu2017high, zou2017random, cheng2022anchor, ding2021object,llerena2021gaussian}. In contrast, SAR sensors enable all-weather, all-day imaging, resilient to cloud cover and illumination changes. However, the inherent speckle noise and low contrast arising from microwave backscatter present significant obstacles to precise localization and classification. Moreover, constructing large-scale SAR datasets \cite{li2024sardet,zhang2025rsar,wu2024fair,li2025saratr,inkawhich20254th,bowald20253rd,ye20253mos} is hindered by the high cost of data acquisition and the laborious fine-grained annotation. Although optical and SAR images exhibit strong complementarity in spatial resolution and environmental robustness, existing single-source \cite{tian2019fcos,zhang2020bridging,sun2024united,WEFT,FSEL,lu2026lwganet} and multi-source object detection methods \cite{he2023multispectral,cao2023multimodal,yuan2022translation,zhang2024e2e} face challenges in addressing cross-domain distribution shifts and suffer from the absence of a standardized evaluation benchmark. To enable robust detection in complex scenarios, there is an urgent need for a large-scale optical-SAR fusion object detection dataset and a standardized benchmarking toolkit.

Currently, a major bottleneck in optical-SAR fusion object detection lies in the lack of large-scale, high-quality, and standardized datasets. As shown in Table \ref{dataset}, most existing SAR datasets are limited to single-source data, suffering from limited scale, narrow category coverage, or inconsistent annotations. For example, HRSID \cite{wei2020hrsid}, SADD \cite{zhang2022sefepnet}, SSDD \cite{zhang2021sar}, SIVED \cite{lin2023sived}, and ShipDataset \cite{wang2019sar} exhibit notable limitations in terms of category diversity and dataset scale. In recent years, the emergence of larger-scale datasets such as SARDet-100K \cite{li2024sardet}, FAIR-CSAR \cite{wu2024fair}, and RSAR \cite{zhang2025rsar} has partially addressed these limitations. However, these datasets are often synthesized by standardization of data from multiple sources, potentially introducing bias during data preprocessing and compromising the fairness of model evaluation. Moreover, constructing paired optical-SAR datasets is prohibitively expensive: acquiring SAR data is already costly, and the acquisition of spatio-temporally aligned optical images is even more difficult, further limiting the construction of multi-source datasets. 

Although OGSOD-1.0 \cite{wang2023category} and OGSOD-2.0 \cite{wang2025cross} represent pioneering efforts in optical-SAR multi-source data, they are primarily designed for cross-modal knowledge distillation. Moreover, their test sets consist solely of SAR images, which limits their applicability for end-to-end multi-source fusion object detection (Other related work can be found in Appendix \ref{Appendix-B}.). To address this issue, we constructed the novel optical-SAR fusion object detection dataset (M4-SAR) covering multi-resolution, multi-polarization, multi-scene, and multi-source using publicly available optical and SAR data provided by Sentinel-1 \cite{torres2012gmes}, and Sentinel-2 \cite{drusch2012sentinel} satellites. To overcome SAR annotation challenges, we propose a semi-supervised optical-assisted labeling strategy that utilizes optical images’ semantic richness to substantially improve annotation quality. As illustrated in Fig. \ref{motivation} and Table \ref{dataset}, M4-SAR includes six key target categories with 112,174 image pairs and 981,862 instances. M4-SAR surpasses existing datasets in instance volume and image numbers, establishing a robust benchmark for multi-source fusion object detection in complex scenarios.

% 图 1 中示例说明如下：
% 1）图像对中光学图像存在云层遮挡，突出展示了 SAR 图像在复杂天气条件下的鲁棒性；
% 2）冬季场景中光学图像被积雪覆盖，目标显著性下降，而 SAR 图像仍清晰可辨；
% 3）低分辨率（60m）光学图像中目标几乎不可识别，而 SAR 图像展现出良好的目标显著性；
% 4）–6）展示了数据集中六类典型目标，尽管部分目标尺寸较小，但依然具备明确的检测价值。

\begin{table*}[t]
\caption{Comparison between existing optical and SAR object detection datasets and the proposed M4-SAR dataset. O and S denote optical and SAR data, respectively.}
\centering
\fontsize{7pt}{7.5pt}\selectfont
\setlength{\tabcolsep}{1.0mm}
\begin{tabular}{l|cccccccc}
\Xhline{1.0pt} 
\textbf{Dataset}  & \textbf{O}  & \textbf{S} & \textbf{Resolution} & \textbf{Image Size}      & \textbf{Cat.}  & \textbf{Instances} & \textbf{Images}   & \textbf{Ins/Img} \\ \hline
HRSC2016 \cite{liu2017high}        & $\checkmark$ & {\color{black!30}\ding{55}} & 0.4M $\sim$ 2M & 300 $\sim$ 1,500  & 1      & 2,976    & 1,061    & 2.80    \\
LEVIR \cite{zou2017random}        & $\checkmark$ & {\color{black!30}\ding{55}} & 0.2M $\sim$ 1M & 600$\times$800  & 3      & 11,028    & 21,952    & 0.50    \\
DIOR-R \cite{cheng2022anchor}        & $\checkmark$ & {\color{black!30}\ding{55}} & 0.5M $\sim$ 30M  & 800$\times$800  & 20      & 190,288    & 23,463    & 8.11    \\
DOTA-v1.0 \cite{ding2021object}        & $\checkmark$ & {\color{black!30}\ding{55}} & - & 800 $\sim$ 4,000  & 15      & 188,282    & 2,806    & 67.10    \\
HRSID \cite{wei2020hrsid}        & {\color{black!30}\ding{55}} & $\checkmark$ & -   & 800$\times$800  & 1      & 16,969    & 5,604    & 3.03    \\
SADD \cite{zhang2022sefepnet}         & {\color{black!30}\ding{55}} & $\checkmark$ & -  & 224$\times$224  & 1      & 7,835     & 883      & 8.87    \\
SSDD \cite{zhang2021sar}         & {\color{black!30}\ding{55}} & $\checkmark$ & -  & 512$\times$512  & 1      & 2,587     & 1,160    & 2.23    \\
SIVED \cite{lin2023sived}        & {\color{black!30}\ding{55}} & $\checkmark$ & -   & 512$\times$512  & 1      & 12,013    & 1,044    & 11.51   \\
ShipDataset \cite{wang2019sar}   & {\color{black!30}\ding{55}} & $\checkmark$ & -  & 256$\times$256  & 1      & 50,885    & 39,729   & 1.28    \\
MSAR \cite{xia2022crtranssar}    & {\color{black!30}\ding{55}} & $\checkmark$ & -  & 256$\times$256  & 4      & 65,202    & 30,158   & 2.16    \\
SARDet-100K \cite{li2024sardet}  & {\color{black!30}\ding{55}} & $\checkmark$ & -   & 512$\times$512  & 6      & 245,653   & 116,598  & 2.11    \\
FAIR-CSAR \cite{wu2024fair}    & {\color{black!30}\ding{55}} & $\checkmark$ & -   & 1,024$\times$1,024  & 5      & 349,002   & 106,672  & 2.11    \\
RSAR \cite{zhang2025rsar}        & {\color{black!30}\ding{55}} & $\checkmark$ & -  & 512$\times$512  & 6      & 183,534   & 95,842   & 1.91    \\
OGSOD-1.0 \cite{wang2023category}    & $\checkmark$ & $\checkmark$ & 20M  & 256$\times$256  & 3      & 48,589    & 18,331   & 2.65    \\
OGSOD-2.0 \cite{wang2025cross}    & $\checkmark$ & $\checkmark$ & 20M   & 256$\times$256  & 3      & 54,000   & 20,359   & 2.65    \\ \hline
\rowcolor{black!5} 
\textbf{M4-SAR (Ours)}                      & $\checkmark$ & $\checkmark$ & \textbf{10M, 60M}  & \textbf{512$\times$512}  & \textbf{6} & \textbf{981,862}   & \textbf{112,174}  & \textbf{8.75}   \\ \Xhline{1.0pt} 
\end{tabular}
\label{dataset}
\end{table*}

Another important challenge is the lack of publicly available algorithms tailored for optical-SAR fusion object detection. Most existing studies are based on single-source datasets or evaluated on limited scenarios, making it difficult to conduct fair performance comparisons. To bridge this gap, we introduce Multi-Source Rotated Object Detector (MSRODet), an open-source toolkit that integrates state-of-the-art fusion detectors (e.g., MHFNet \cite{zhang2026mhfnet}, CFT \cite{qingyun2022cross}, CLANet \cite{he2023multispectral}, CSSA \cite{cao2023multimodal}, CMADet \cite{song2024misaligned}, ICAFusion \cite{shen2024icafusion}, and MMIDet \cite{zeng2024mmi}), providing a standardized evaluation framework for researchers utilizing the M4-SAR dataset. MSRODet not only enables fair and reproducible benchmarking across different fusion methods but also lays a solid foundation for the development of future multi-source fusion object detection algorithms.

Moreover, extensive experiments demonstrate that the intrinsic differences in the imaging mechanisms of optical and SAR sensors lead to significant domain gaps (Fig. \ref{overall-network} (a)), thereby constraining the performance of optical-SAR fusion object detection. To address this issue, we propose the first end-to-end framework for optical-SAR fusion object detection, termed E2E-OSDet. This framework effectively fuses features from both modalities, alleviates cross-modal domain gaps, and results in significant improvements in detection performance.

% This work offers substantial contributions in terms of dataset construction, toolkit development, and algorithm design. 
The main contributions of this work are summarized as follows:
\begin{itemize}
\item \textbf{A Novel Optical-SAR Dataset:} M4-SAR is a novel optical-SAR fusion object detection dataset with multi-resolution, multi-polarization, and multi-scene characteristics, containing 112,174 image pairs and 981,862 instances. This dataset comprehensively covers complex scenes and diverse target categories frequently encountered in multi-source fusion detection tasks, thereby providing a robust foundation for advancing research in this field.
\item \textbf{Multi-source Evaluation Toolkit (MSRODet) and E2E-OSDet:} We construct MSRODet, a comprehensive benchmarking toolkit incorporating state-of-the-art fusion object detection methods tailored for optical-SAR fusion object detection, which provides a standardized and reproducible evaluation platform based on the M4-SAR dataset. In addition, we propose E2E-OSDet, a novel end-to-end optical-SAR fusion object detection framework, which effectively mitigates significant cross-domain discrepancies between optical and SAR modalities at the feature level.
\end{itemize}

\begin{figure*}[t]	
    \centering
    \includegraphics[width=0.99\linewidth]{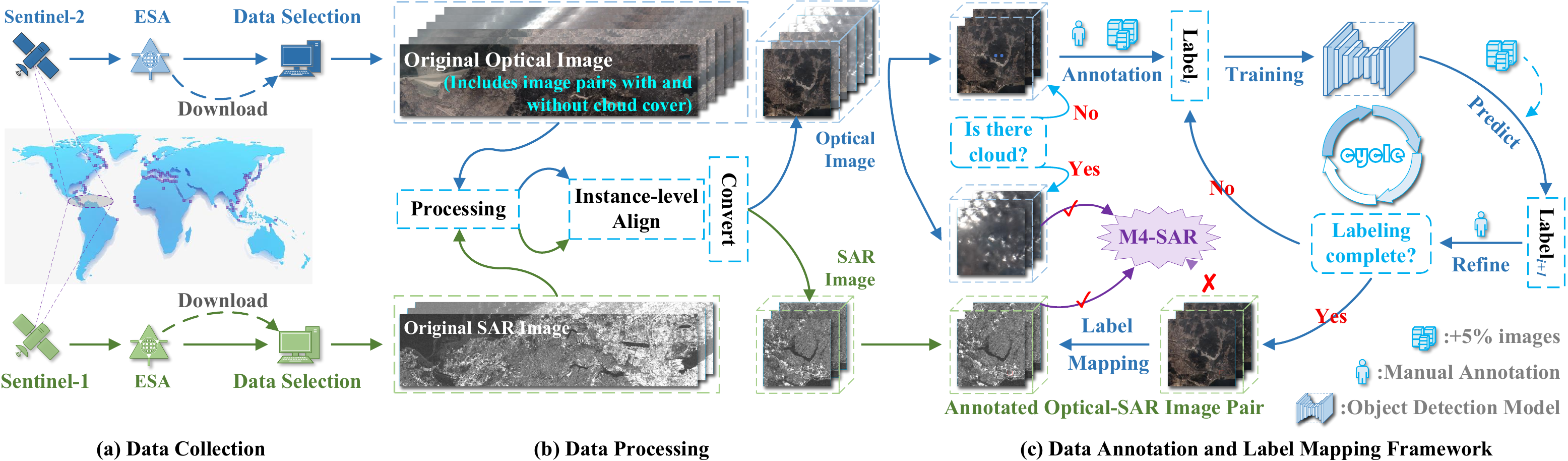}
    \caption {Pipeline for M4-SAR dataset construction.
    }
    \label{data-process}
\end{figure*}

\section{M4-SAR Dataset Construction}
\label{sec-3}
\subsection{Overview of the Proposed Dataset Construction Pipeline} 
Fig. \ref{data-process} illustrates the overall pipeline of the proposed dataset construction process. In the first stage, we acquire optical and SAR images from Sentinel-1 and Sentinel-2 satellites provided by the European Space Agency (ESA). In the second stage, the acquired images are pre-processed and aligned. After completing the alignment, we crop aligned image pairs into fixed-size patches to generate initial unlabeled samples. In the third stage, we introduce a semi-supervised annotation strategy guided by optical images to ensure high-quality labeling.

\subsection{Data Collection} 
In Fig. \ref{data-process} (a), we collected the required data via the open-access platform provided by the ESA. A map-based selection tool was used to specify areas of interest, covering representative coastal cities, including Tianjin, Los Angeles, Hong Kong, and London. We then retrieved SAR images along with their corresponding digital elevation model (DEM) data according to specific criteria. For optical imagery, Sentinel-2 images covering the same regions as the Sentinel-1 data were acquired, specifically including the red (R), green (G), and blue (B) spectral bands. To minimize variations caused by temporal discrepancies between the two modalities, we selected image pairs with a time interval of less than 10 days. This setting ensures that targets of interest (e.g., bridges, oil tanks) maintain structural consistency within the short interval, thereby enhancing the accuracy of subsequent instance-level alignment. The product type for optical images is L2A, with cloud coverage limited to 0–10\% (the corresponding cloud-free (0\%) images are used for high-quality annotation) to guarantee high visual fidelity. The SAR images are provided in the Single Look Complex (SLC) format to preserve full coherence information. The collected data span the period from 2020 to 2022 and encompass multiple spatial resolutions and polarization modes. The optical images have 10M, 20M, and 60M resolutions, and the SAR data cover both vertical-horizontal (VH) and vertical-vertical (VV) polarizations.

\subsection{Data Processing}
As shown in Fig. \ref{data-process} (b), we combined the red, green, and blue channels to generate standard optical images using data processing software (ENVI). The optical images were provided in three spatial resolutions: 10M, 20M, and 60M. To highlight the complementary characteristics of optical and SAR images, we retained only the high-resolution (10M) and low-resolution (60M) variants. For the SAR data, we first applied multi-look averaging to the SLC images in both the azimuth and range directions to reduce speckle effects and improve radiometric resolution. Subsequently, filtering techniques were employed to suppress speckle noise. Using the corresponding DEM data, we geocoded and radiometrically calibrated the SAR images and conducted geometric alignment and correction to improve spatial localization precision. After preprocessing, the optical and SAR images were aligned based on their geographic coordinates. As the overlap between the two modalities was partial, the aligned image pairs were cropped into 1,024$\times$1,024 patches, and only those with valid overlapping regions were retained for subsequent annotation. Notably, this initial cropping was intended to facilitate efficient manual annotation. After annotation was completed, we further partitioned the annotated images into non-overlapping 512$\times$512 patches to meet the demands of different downstream applications.

\subsection{Data Annotation and Label Mapping}
As the acquisition times of Sentinel-1 \cite{torres2012gmes} and Sentinel-2 \cite{drusch2012sentinel} are not perfectly synchronized, instance-level alignment of dynamic targets (e.g., ships) becomes difficult, which compromises the accuracy of label transfer. To address this issue, we focus on static and semantically meaningful targets, such as bridges, harbors, oil tanks, playgrounds, airports, and wind turbines, and annotate them using oriented bounding boxes. SAR's unique imaging characteristics cause ambiguous target boundaries, challenging accurate annotation by non-experts. In contrast, optical images (cloudless) provide clearer visual cues that facilitate both target identification and annotation transfer. With precise alignment, annotations derived from optical images can be reliably transferred to SAR images, thereby significantly reducing manual annotation costs.

Given the scale of the dataset, fully manual labeling is both time-consuming and labor-intensive. To address this challenge, we propose a semi-supervised annotation strategy, as illustrated in Fig. \ref{data-process} (c). The procedure involves the following steps: First, 5\% of the entire cloudless optical images are manually annotated, ensuring that each selected image contains all six target categories: bridges, harbors, oil tanks, playgrounds, airports, and wind turbines. This subset establishes a high-quality foundation for subsequent annotation. Once manually labeled, a detector \cite{Jocher_Ultralytics_YOLO_2023} is trained using this subset (with the same data used for both training and evaluation during the initial phase). The trained model is then employed to generate pseudo-labels for an additional 5\% of the data, which are subsequently refined through manual correction and added to the labeled set. This updated set is used to retrain the detector in the next iteration. The annotation process is repeated iteratively until the full dataset is labeled. During this procedure, images without valid target instances are excluded to enhance annotation efficiency of the M4-SAR dataset. 

During the annotation process, we use cloud-free images to ensure annotation accuracy. After annotation is complete, the cloud-free images are replaced with the corresponding cloud-containing images. Based on instance-level alignment, we achieve high-quality label transfer, enabling accurate projection of annotation information from optical images to their corresponding SAR counterparts. As a result, we construct M4-SAR, the large-scale dataset for optical-SAR fusion object detection, which offers a solid foundation for subsequent method evaluation and algorithm development. More dataset details are presented in Appendix~\ref{Appendix-A}.

\subsection{Data Analysis}
As shown in Table \ref{dataset}, our proposed M4-SAR dataset comprises 112,174 image pairs, 981,862 instances, and six representative target categories: bridges, harbors, oil tanks, playgrounds, airports, and wind turbines. Each image has a resolution of 512$\times$512 pixels and contains an average of 8.75 instances, highlighting the dense object distribution typical in remote sensing scenes. As illustrated in Fig. \ref{motivation} (c), rotated bounding boxes offer significantly higher localization accuracy compared to horizontal bounding boxes. The challenge analysis of category-specific attributes in the M4-SAR dataset is presented as follows:

\begin{figure*}[!t]	
    \centering
    \includegraphics[width=1.0\linewidth]{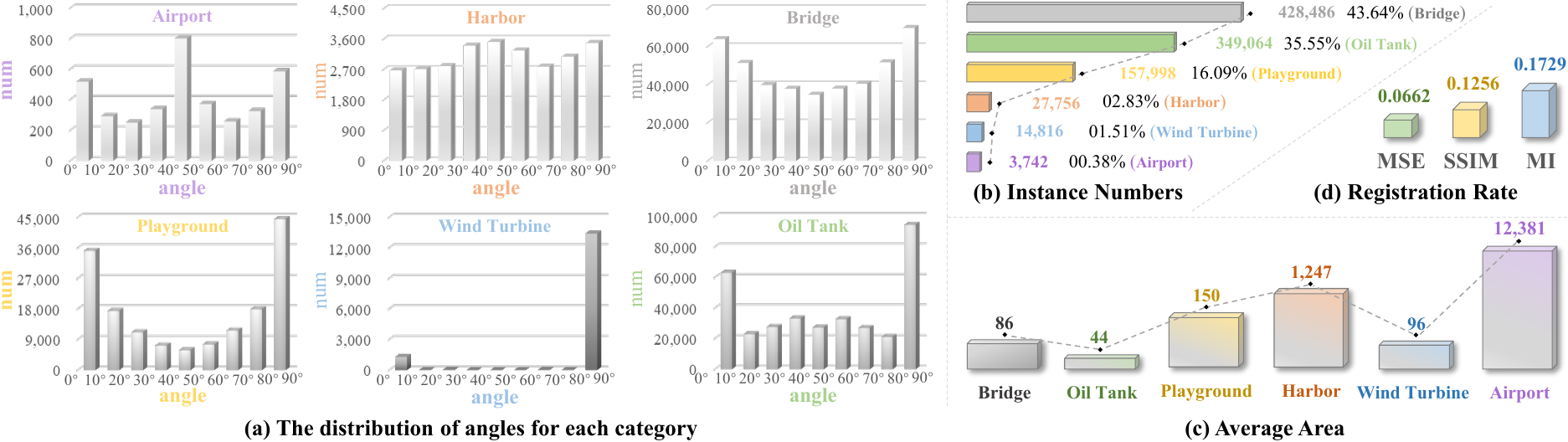}
    \caption {Statistical visualization of category attributes in the proposed M4-SAR dataset. (a) Angle distribution of instances in each category. (b) Percentage of instances per category. (c) Average pixel area of instances per category.
    }
    \label{data-statistics}
\end{figure*}

\begin{itemize}
    \item \textbf{Aspect Ratio Challenge:} The aspect ratio of an object plays a critical role in detection accuracy. Fig. \ref{motivation} (b) illustrates the distribution of aspect ratios across categories. When the aspect ratio approaches 1, targets tend to appear square, potentially leading to recognition ambiguities. Conversely, extremely large aspect ratios increase the model’s sensitivity to object shapes, thus posing additional challenges to accurate detection.
    \item \textbf{Angle Diversity Challenge:} All target orientation angles are normalized to $\theta$ ($\theta \in [0, \pi/2)$, consistent with YOLO-OBB \cite{Jocher_Ultralytics_YOLO_2023} rules.), and their distributions are visualized in Fig. \ref{data-statistics} (a). Notably, offshore wind turbines tend to exhibit square-like appearances, with orientation angles predominantly distributed to 0$^{\circ}$ and 90$^{\circ}$ due to the absence of surrounding reference structures and the top-down perspective of satellite imagery. Although oil tanks are circular, their labeled angles are contextually defined based on surrounding spatial features. This highlights the importance of contextual cues in remote sensing images and reflects the task's sensitivity.
    \item \textbf{Category Imbalance Challenge:} Fig. \ref{data-statistics} (b) displays the distribution of instances by category, revealing that bridges (43.64\%) constitute the largest proportions in the dataset. This imbalance reflects the actual occurrence frequency of typical targets in coastal regions: static infrastructure such as bridges and oil tanks dominates satellite imagery. The relatively low number of harbor (2.83\%) instances is primarily due to their extensive spatial coverage, with a single port typically corresponding to a single instance.
    \item \textbf{Scale Inconsistency Challenge:} Fig. \ref{data-statistics} (c) illustrates the average pixel area of each category, highlighting the substantial variation in object scales across different categories. Such large discrepancies in object size (i.e., object area) introduce a critical and commonly encountered challenge for robust object detection, as models must simultaneously handle extremely small and large targets within a unified framework.
    \item \textbf{Modal Weak Alignment Challenge:} The optical and SAR images are aligned only using geographic coordinates, resulting in coarse rather than pixel-level correspondence. Together with non-simultaneous acquisitions and dynamic scene changes, this leads to notable cross-modal misalignment, as reflected by the low Mean Squared Error (MSE: 0.0662), Structural Similarity Index Measure (SSIM: 0.1256), and Mutual Information (MI: 0.1729) scores shown in Fig. \ref{data-statistics} (d). This weak alignment poses a significant challenge for multi-source fusion object detection.
    % , requiring models to be robust to spatial misalignment and cross-modal inconsistency.
\end{itemize}

\begin{figure*}[!t]	
    \centering
    \includegraphics[width=1.0\linewidth]{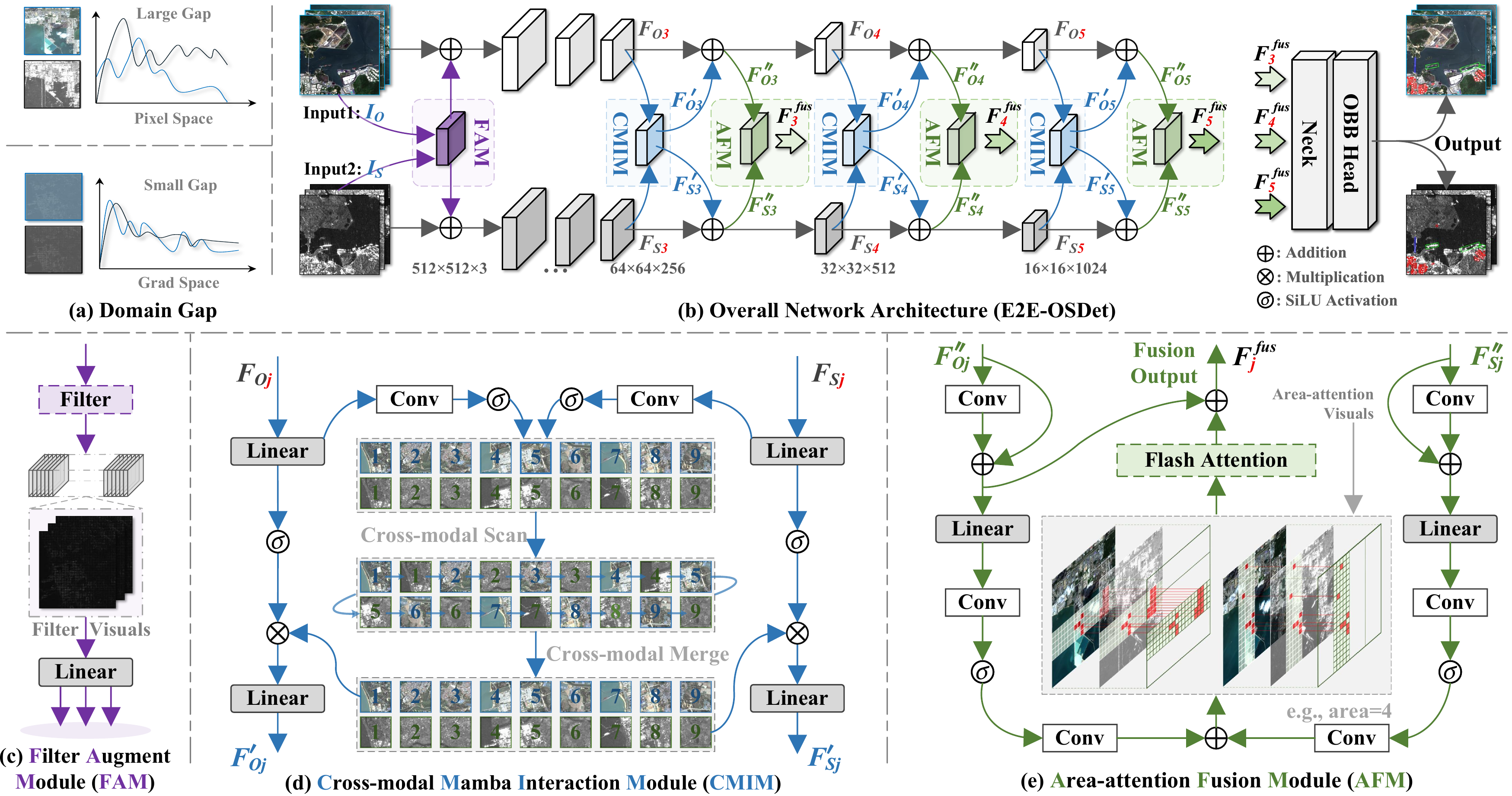}
    \caption {(a) Domain gap between optical and SAR images. (b) Overall framework of the proposed end-to-end optical-SAR fusion object detection (E2E-OSDet). Architectural details of the proposed filter augment module (c), cross-modal Mamba interaction module (d), and area-attention fusion module (e).
    }
    \label{overall-network}
\end{figure*}

\section{Evaluation Tools and Methods}
\label{sec-4}

\subsection{Optical-SAR Fusion Object Detection Tools (MSRODet)}
\label{4-1}
Due to the lack of publicly available methods specifically designed for optical-SAR fusion object detection, conducting a fair performance comparison remains challenging. To address this limitation, we propose a Multi-Source Rotated Detector (MSRODet) built upon existing multi-source detection frameworks (e.g., RGB-infrared, RGB-depth). MSRODet integrates several representative methods, including MHFNet \cite{zhang2026mhfnet}, CFT \cite{qingyun2022cross}, CLANet \cite{he2023multispectral}, CSSA \cite{cao2023multimodal}, CMADet \cite{song2024misaligned}, ICAFusion \cite{shen2024icafusion}, and MMIDet \cite{zeng2024mmi}, thereby providing a standardized and accessible benchmark for evaluating optical-SAR fusion object detection approaches. As shown in Fig.~\ref{overall-network} (a), empirical analysis reveals that significant domain gaps remain a primary obstacle to effective optical-SAR fusion. These discrepancies primarily arise from fundamental differences in visual characteristics and imaging mechanisms between optical and SAR modalities. Consequently, designing detection frameworks specifically tailored to optical-SAR fusion is essential for enhancing overall detection performance.

\subsection{End-to-End Optical-SAR Fusion Object Detection (E2E-OSDet)}
Multi-source fusion object detection remains underexplored in the context of optical-SAR fusion, and significant limitations persist in addressing domain discrepancies between optical and SAR data. As shown in Fig.~\ref{overall-network} (b), to address these challenges, we propose a novel end-to-end optical-SAR fusion object detection framework, termed E2E-OSDet, which systematically addresses cross-modal challenges across three levels: data input, domain alignment, and feature fusion.

\textbf{Filter Augment Module (FAM):} First, to reduce the domain gap between optical and SAR images, we propose a Filter Augment Module (FAM, Fig.~\ref{overall-network} (c)) leveraging classical image filtering operators (e.g., HOG~\cite{dalal2005histograms}, Canny~\cite{canny1986computational}, Haar~\cite{viola2001rapid}, Grad~\cite{li2024predicting}, and WST~\cite{mallat2012group}). This module transforms the low-dimensional and sparse SAR representations into a more discriminative high-dimensional space, effectively mitigating the impact of domain differences.

\textbf{Cross-modal Mamba Interaction Module (CMIM):} Second, most existing methods focus on spatial-level fusion but often neglect the discrepancies in the feature domain. Mamba~\cite{gu2024mamba} possesses strong state-space modeling capabilities, which enables effective cross-modal sequence-level interactions and alleviates feature domain bias. However, current Mamba-based methods~\cite{zhan2025mambasod,dong2025fusion,hu2025exploiting} are hindered by complex architectures and intricate interaction mechanisms, limiting their practical applicability. To address this, we propose a Cross-modal Mamba Interaction Module (CMIM, Fig.~\ref{overall-network} (d)), which enables deep cross-modal interactions through simple and efficient operations.

\textbf{Area-Attention Fusion Module (AFM):} Finally, to further enhance the discriminability of fused features across different spatial regions, we introduce an Area-Attention Fusion Module (AFM). This module captures salient features of local targets and strengthens responses in critical regions, thereby enhancing the effectiveness of feature fusion. The detailed structure of AFM is illustrated in Fig.~\ref{overall-network} (e). For additional E2E-OSDet details, please refer to Appendix~\ref{Appendix-C}.

\begin{table*}[!t]
\caption{Quantitative results on the M4-SAR. O and S denote optical and SAR data, respectively. Param. denotes the parameter count of the model. Inf.T indicates inference time per image. The best two results are highlighted in \textbf{Black} and \underline{mark}.}
\centering
\fontsize{7pt}{7.5pt}\selectfont
\setlength{\tabcolsep}{0.4mm}
\begin{tabular}{l|cc|c|c|c|ccccccc|cc}
\Xhline{1.0pt} 
\multirow{2}{*}{\textbf{Method}} &
  \multirow{2}{*}{\textbf{O}} &
  \multirow{2}{*}{\textbf{S}} &
  \textbf{Param.} &\textbf{FLOPs} &
  \textbf{Inf.T} &
  \multicolumn{7}{c|}{\textbf{AP}$_{50}$ \small{$\uparrow$}} &
  \textbf{AP}$_{75}$ &
  \textbf{mAP} \\ \cline{7-13}
 &
   &
   &
  \textbf{(M)} \small$\downarrow$&
  \textbf{(G)} \small$\downarrow$&
  \textbf{(ms)} \small$\downarrow$&
  {Bri.} &
  \multicolumn{1}{c}{{Har.}} &
  \multicolumn{1}{c}{{Oil.}} &
  \multicolumn{1}{c}{{Ply.}} &
  \multicolumn{1}{c}{{Apr.}} &
  \multicolumn{1}{c}{{Win.}} &
  \multicolumn{1}{c|}{\textit{ALL}} &
  \small{$\uparrow$} &
  \small{$\uparrow$} \\ \hline
R-FCOS  \cite{tian2019fcos}                       & $\checkmark$ & {\color{black!30}\ding{55}} & 31.9&51.6& 17.6&32.8 &79.8 &29.7 &67.0 &78.9 &54.8 & 57.2 & 26.4 & 30.2 \\
R-ATSS  \cite{zhang2020bridging}                  & $\checkmark$ & {\color{black!30}\ding{55}} & 36.0&51.9& 19.1&33.2 &80.0 &29.3 &66.6 &78.3 &54.8 & 57.0 & 27.1 & 30.0 \\
O-RepPoint  \cite{li2022oriented}                 & $\checkmark$ & {\color{black!30}\ding{55}} & 36.6&48.6& 16.1&39.7 &71.9 &21.2 &56.3 &65.5 &53.1 & 51.3 & 17.9 & 26.7 \\
RTMDet  \cite{lyu2022rtmdet}                      & $\checkmark$ & {\color{black!30}\ding{55}} & 8.9 &51.0& 11.1&34.0 &79.8 &30.3&61.1 &79.1 &55.9 & 56.7 & 26.5 & 29.2 \\
PSC  \cite{yu2023phase}                           & $\checkmark$ & {\color{black!30}\ding{55}} & 31.9&51.7& 18.3&30.7 &79.7 &25.9 &57.4 &78.1 &49.4 & 53.5 & 26.2 & 28.5 \\
LSKNet  \cite{li2024lsknet}                       & $\checkmark$ & {\color{black!30}\ding{55}} & 21.8&43.0& 20.4&42.9 &80.6 &34.9&78.4 &79.1 &54.0 & 61.6 & 31.2 & 34.0 \\
YOLOv5  \cite{Jocher_Ultralytics_YOLO_2023}       & $\checkmark$ & {\color{black!30}\ding{55}} & 8.1&12.8& 7.2&61.4 &88.0 &56.6 &76.4 &91.3 &91.8 & 77.6 & 59.8 & 52.9 \\
YOLOv6  \cite{li2022yolov6}                       & $\checkmark$ & {\color{black!30}\ding{55}}   & 16.0&27.7& 6.1&57.6 &82.1 &52.7 &67.6 &89.4 &90.4 & 73.3 & 64.9 & 48.8 \\
YOLOv8  \cite{Jocher_Ultralytics_YOLO_2023}       & $\checkmark$ & {\color{black!30}\ding{55}}   & 10.0&15.7& 6.7 &63.6 &88.6 &57.9 &80.0 &91.3 &92.5 & 79.0 & 61.6 & 54.4 \\
YOLOv9  \cite{wang2024yolov9}                     & $\checkmark$ &  {\color{black!30}\ding{55}}  & 6.4&15.1& 14.4&61.9 &87.7 &55.3 &74.4 &90.0 &92.4 & 76.9 & 58.9 & 52.5 \\
YOLOv10  \cite{wang2024yolov10}                   & $\checkmark$ &  {\color{black!30}\ding{55}}  & 7.5&14.5& 7.8&65.2 &89.9 &58.8 &82.3 &90.8 &93.5 & 80.1 & 62.8 & 55.7 \\
YOLOv11  \cite{Jocher_Ultralytics_YOLO_2023}      & $\checkmark$ &  {\color{black!30}\ding{55}}  & 9.7&14.4& 6.5&64.3 &90.0 &57.8 &80.7 &90.9 &91.5 & 79.2 & 60.5 & 54.3 \\
YOLOv12  \cite{tian2025yolov12}                       & $\checkmark$ & {\color{black!30}\ding{55}} & 9.4&14.4&  13.4& 62.7 &  88.6   &  57.1 &  80.1 & 91.0 &  92.7 & 78.7 & 61.1 & 54.0\\
YOLO26  \cite{Jocher_Ultralytics_YOLO_2023}       & $\checkmark$ & {\color{black!30}\ding{55}}   & 10.5&15.7& 9.1&63.0 &88.2 &58.3 &80.6 &90.3 &90.1 & 78.4 & 62.8 & 55.4 \\
YOLO-Master  \cite{lin2025yolo}       & $\checkmark$ & {\color{black!30}\ding{55}}   & 29.3&35.6& 15.7&64.4 &90.0 &59.0 &80.7 &92.3 &93.6 & 80.0 & 63.9 & 56.3\\
\hline
R-FCOS  \cite{tian2019fcos}                       & {\color{black!30}\ding{55}}& $\checkmark$  & 31.9&51.6& 17.6&23.9 &80.1 &27.9 &60.1 &75.8 &43.5 & 51.9 & 26.4 & 28.3 \\
R-ATSS  \cite{zhang2020bridging}                  &{\color{black!30}\ding{55}} & $\checkmark$  & 36.0&51.9& 19.1&25.3 &80.3 &26.8 &62.3 &77.0 &43.9 & 52.6 & 24.9 & 27.8 \\
O-RepPoint  \cite{li2022oriented}                 & {\color{black!30}\ding{55}} & $\checkmark$ & 36.6&48.6& 16.1&26.2 &78.6 &21.9 &53.7 &59.9 &48.8 & 48.2 & 13.9 & 20.4 \\
RTMDet  \cite{lyu2022rtmdet}                      & {\color{black!30}\ding{55}} & $\checkmark$ & 8.9&51.0& 11.1&16.2 &71.0 &21.1 &39.3 &68.9 &39.6 & 42.7 & 20.9 & 22.3\\
PSC  \cite{yu2023phase}                           & {\color{black!30}\ding{55}} & $\checkmark$ & 31.9&51.7& 18.3&22.3 &80.0 &20.6 &51.9 &75.7 &39.4 & 48.3 & 24.5 & 26.5 \\
LSKNet  \cite{li2024lsknet}                       & {\color{black!30}\ding{55}} & $\checkmark$ & 21.8&43.0& 20.4&33.4 &80.2 &28.4 &69.3 &76.8 &51.2 & 56.6 & 27.1 & 30.4 \\
YOLOv5  \cite{Jocher_Ultralytics_YOLO_2023}       &{\color{black!30}\ding{55}} &  $\checkmark$  & 8.1&12.8& 7.2&40.5 &85.3 &41.0 &64.3 &85.5 &89.9 & 67.7 & 44.0 & 41.3 \\
YOLOv6  \cite{li2022yolov6}                       & {\color{black!30}\ding{55}} &  $\checkmark$  & 16.0&27.7& 6.1&36.5 &79.4 &32.8 &53.5 &85.7 &86.7 & 62.4 & 37.6 & 36.5 \\
YOLOv8  \cite{Jocher_Ultralytics_YOLO_2023}       & {\color{black!30}\ding{55}} &  $\checkmark$  & 10.0&15.7& 6.7&44.9 &85.8 &42.5 &67.9 &87.5 &89.2 & 69.6 & 46.1 & 43.2 \\
YOLOv9  \cite{wang2024yolov9}                     & {\color{black!30}\ding{55}}&  $\checkmark$  & 6.4&15.1& 14.4&43.9 &84.8 &38.4 &61.5 &87.9 &88.8 & 67.6 & 44.2 & 41.4 \\
YOLOv10  \cite{wang2024yolov10}                   & {\color{black!30}\ding{55}} & $\checkmark$  & 7.5&14.5& 7.8&49.8 &87.0 &45.5 &73.4 &87.9 &90.4 & 72.3 & 50.8 & 46.2 \\
YOLOv11  \cite{Jocher_Ultralytics_YOLO_2023}      & {\color{black!30}\ding{55}} &  $\checkmark$ & 9.7&14.4& 6.5&46.9 &87.2 &44.6 &71.1 &88.8 &89.2 & 71.3 & 48.6 & 44.7 \\
YOLOv12  \cite{tian2025yolov12}                   & {\color{black!30}\ding{55}} & $\checkmark$ & 9.4&14.4&  13.0&  46.4  &  87.7  &  43.6  &  70.8 & 90.0 & 90.7 & 71.5 & 49.2 &  45.4 \\
YOLO26  \cite{Jocher_Ultralytics_YOLO_2023}       &  {\color{black!30}\ding{55}}   &$\checkmark$& 10.5&15.7& 9.1   & 48.8& 88.3& 48.9& 75.8& 88.9& 88.9& 73.3& 53.8& 48.2 \\
YOLO-Master  \cite{lin2025yolo}       &  {\color{black!30}\ding{55}}   &$\checkmark$& 29.3&35.6& 15.7   & 49.4& 89.3& 48.3& 75.1& 90.6& 90.8& 73.9& 53.1& 48.5 \\
\hline
COMO \cite{liu2026cross}                       & $\checkmark$ &   $\checkmark$  &22.6&24.3&37.1&72.3 & 90.8 &60.4 &88.7 &90.5 &96.3 & 83.2 & 66.3& 57.9 \\
MHFNet \cite{zhang2026mhfnet}                       & $\checkmark$ &   $\checkmark$  &16.1&26.7&19.2&73.7 & 91.4 &\underline{61.8} &89.9 &90.4 &95.6 & 83.8 & 65.6& 58.0 \\
CFT \cite{qingyun2022cross}                       & $\checkmark$ &   $\checkmark$    &53.8&31.8& 40.6&\underline{75.8} &\underline{92.5} &61.3 &91.6 &90.3 &96.3 & 84.6 & \underline{68.9} & \underline{59.9} \\
CLANet \cite{he2023multispectral}                 & $\checkmark$ &  $\checkmark$ &48.2&53.4&29.1&74.8 &92.2 &60.7 &91.3 & 91.6 &\underline{97.2} & 84.6 & 68.5 & 59.6 \\
CSSA \cite{cao2023multimodal}                     & $\checkmark$ &   $\checkmark$  &13.5&21.4& 12.3&73.3 &91.7 &59.3 &88.9 &91.6 &95.8 & 83.4 & 66.4 & 58.0 \\
CMADet \cite{song2024misaligned}                 & $\checkmark$ &  $\checkmark$ &41.5&31.8&46.7&70.9 &90.7 &52.0 &86.4 &91.7 &97.1 & 81.5 & 63.5 & 55.7 \\
ICAFusion \cite{shen2024icafusion}                & $\checkmark$ & $\checkmark$ &29.0&26.7&23.6&74.7 &91.9 &60.9 &91.0 &\underline{91.8} &96.7 & 84.5 & 67.3 & 58.8 \\
MMIDet \cite{zeng2024mmi}                         & $\checkmark$ &  $\checkmark$    &53.8&31.8&41.9&74.9 & \textbf{92.6} &61.1 &\underline{91.7} &91.4 &97.0 & \underline{84.8} & 68.6 & 59.8 \\
\hline
\rowcolor{black!5}
\textbf{E2E-OSDet}                                & $\checkmark$ &   $\checkmark$          &24.7&40.7&20.9& \textbf{77.7} &90.7 & \textbf{64.3} & \textbf{91.8} & \textbf{92.1} & \textbf{97.8} &  \textbf{85.7} &  \textbf{70.3} &  \textbf{61.4} \\ \Xhline{1.0pt} 
\end{tabular}
\label{all-detection}
\end{table*}

\section{Experiments and Analysis} 
\label{sec-5}
\subsection{Experiment Setup and Implementation Details}
Experiments were conducted on two NVIDIA RTX 4090 GPUs. All fusion models (MSRODet, COMO, and the proposed E2E-OSDet) were trained over 300 epochs with the SGD optimizer, using a learning rate of 0.01, momentum of 0.937, and weight decay of 5$\times$10$^{-4}$. The batch size was set to 64, and the input images were resized to a resolution of 512$\times$512. To ensure fair performance evaluation, all methods were evaluated using a unified framework comprising the YOLOv11 \cite{Jocher_Ultralytics_YOLO_2023} backbone and the oriented detection head from YOLOv8 \cite{Jocher_Ultralytics_YOLO_2023}. These fusion models were trained from scratch, without leveraging any pre-trained weights. During evaluation, we adopted the standard COCO evaluation protocol to measure Average Precision (AP), including $AP_{50}$, $AP_{75}$, and $mAP$.

\begin{figure*}[!t]	
    \centering
    \includegraphics[width=1.0\linewidth]{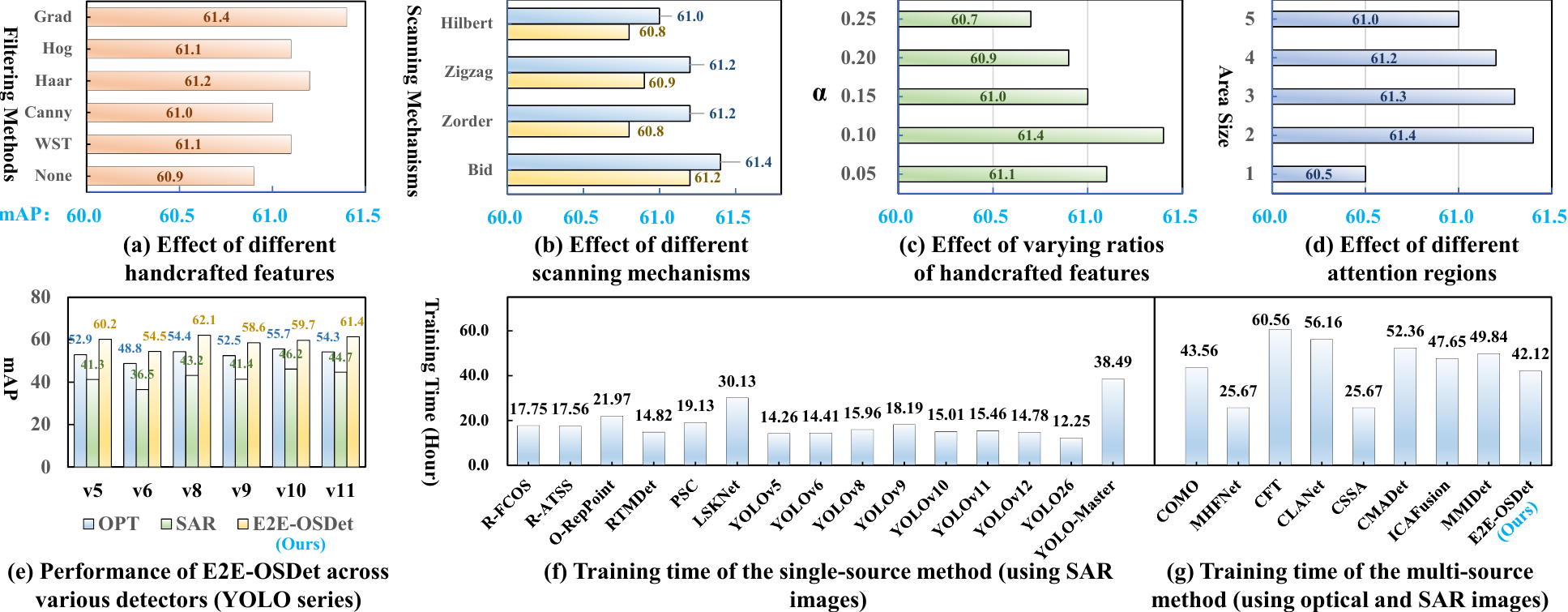}
    \caption {Comprehensive performance evaluation under different parameter settings.  
    }
    \label{ablation-experiment}
\end{figure*}

% \subsection{Quantitative Evaluation} 
\subsection{Main Results} 
% We first compared multi-source fusion methods (such as MHFNet, CFT, CLANet, CSSA, CMADet, ICAFusion, MMIDet, and E2E-MFD) with single-source detection methods (such as R-FCOS, R-ATSS, Roi-Trans, O-RepPoint, RTMDet, and the YOLO series). 
Experimental results (see Table \ref{all-detection}) indicate that detection accuracy using SAR images alone is significantly lower than when using optical images alone. This gap primarily stems from the unique imaging mechanism of SAR: even when targets are visible in the image, SAR struggles to accurately estimate the shape of target bounding boxes. In contrast, all multi-source fusion methods outperform their corresponding single-modal baselines, clearly demonstrating that cross-modal fusion effectively leverages the complementarity between SAR and optical data to significantly reduce false positives and enhance detection accuracy. Among these, our proposed E2E-OSDet achieves the highest performance across all fusion methods, with a $mAP$ of 61.4\%, fully validating the importance of mitigating domain differences in cross-modal detection. Furthermore, E2E-OSDet exhibits high efficiency with only 24.7M parameters and achieves an inference speed of 20.9ms. As shown in Fig. \ref{ablation-experiment} (e), we also evaluated its generalization capability across different detection frameworks (YOLO series, including v5, v6, v8, v9, v10.), demonstrating strong transferability and robustness. 

% We compared multi-source fusion methods (e.g., CFT, CLANet, CSSA, CMADet, ICAFusion, MMIDet, and E2E-MFD) with single-source methods (e.g., R-FCOS, R-ATSS, Roi-Trans, O-RepPoint, RTMDet, PSC, KFIoU, LSKNet, and YOLO series) that utilize either optical or SAR data exclusively. As shown in Table \ref{all-detection}, the detection accuracy achieved using SAR data alone is considerably lower than that of the optical-only counterpart. This performance gap is primarily attributed to the unique imaging principles of SAR, which hinder the accurate estimation of bounding box shapes, even when targets are visually identifiable. All multi-source fusion methods outperform their corresponding unimodal baselines, demonstrating the effectiveness of cross-modal fusion in leveraging SAR data to complement optical information (and vice versa), while significantly reducing false positives. Notably, our proposed method, E2E-OSDet, achieves the highest performance among all fusion-based approaches, reaching an $mAP$ of 61.4\%, which underscores the importance of mitigating domain discrepancies to improve detection accuracy. Furthermore, E2E-OSDet is also highly efficient, with only 24.7M parameters and an inference time of 20.9ms. Finally, we evaluate the generalizability of E2E-OSDet across various detection frameworks, with the results presented in Fig. \ref{ablation-experiment} (e).

\begin{table}[!t]
\caption{Ablation analysis of E2E-OSDet on M4-SAR dataset.}
\centering
\fontsize{7.0pt}{7.5pt}\selectfont
\setlength{\tabcolsep}{1.6mm}
\begin{tabular}{c|ccc|lccc}
\Xhline{1.0pt} 
$\textbf{No.}$ & $\textbf{FAM}$ & $\textbf{CMIM}$ & $\textbf{AFM}$ & \textbf{Train-Params.}&$\textbf{AP}_{50}$\small{ $\uparrow$}  & $\textbf{AP}_{75}$\small{ $\uparrow$}  & $\textbf{mAP}$\small{ $\uparrow$}  \\
\hline
(a) & {\color{black!30}\ding{55}} & {\color{black!30}\ding{55}} & {\color{black!30}\ding{55}} &14.17M {\color{black!60}Baseline}& 83.4 & 66.4 & 58.0 \\
(b) &$\checkmark$& {\color{black!30}\ding{55}} & {\color{black!30}\ding{55}} &14.17M {\color{black!60}+0M}& 84.5 & 67.1 & 58.9 \\
(c) &{\color{black!30}\ding{55}}&$\checkmark$&{\color{black!30}\ding{55}}& 19.01M {\color{black!60}+5.68M}& 84.9 & 68.6 & 59.6 \\
(d) &{\color{black!30}\ding{55}}&{\color{black!30}\ding{55}}&$\checkmark$& 19.84M {\color{black!60}+4.84M}& 84.6 & 68.9 & 60.0 \\
(e) &$\checkmark$&$\checkmark$&{\color{black!30}\ding{55}}&19.84M {\color{black!60}+4.84M}& 85.5 & 69.9 & 61.0\\  % no
(f) &$\checkmark$&{\color{black!30}\ding{55}}&$\checkmark$&19.01M {\color{black!60}+5.68M}& 84.9 & 69.5 & 60.7\\  % no
(g) &{\color{black!30}\ding{55}}&$\checkmark$&$\checkmark$&24.69M {\color{black!60}+10.52M}& 85.2 & 70.0 & 60.9 \\
\rowcolor{black!5}
(h) &$\checkmark$&$\checkmark$&$\checkmark$&24.69M {\color{black!60}+10.52M}& \textbf{85.7} &  \textbf{70.3} &  \textbf{61.4} \\
\Xhline{1.0pt} 
\end{tabular}
\label{albation}
\end{table}

\begin{figure}[!t]	
    \centering
    \includegraphics[width=0.7\linewidth]{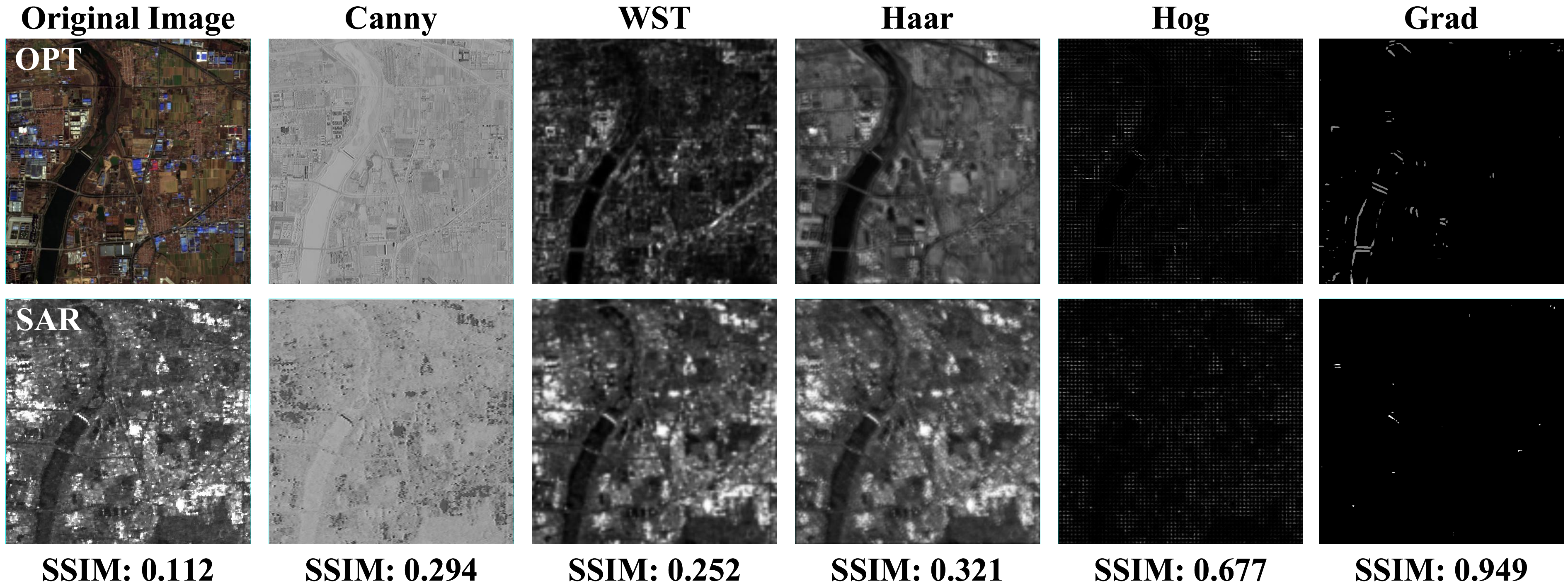}
    \caption {Visualization results of handcrafted features (Canny, WST, Haar, Hog, Grad).
    }
    \label{vis-filter-1}
\end{figure}

\begin{figure*}[!t]	
    \centering
    \includegraphics[width=1.0\linewidth]{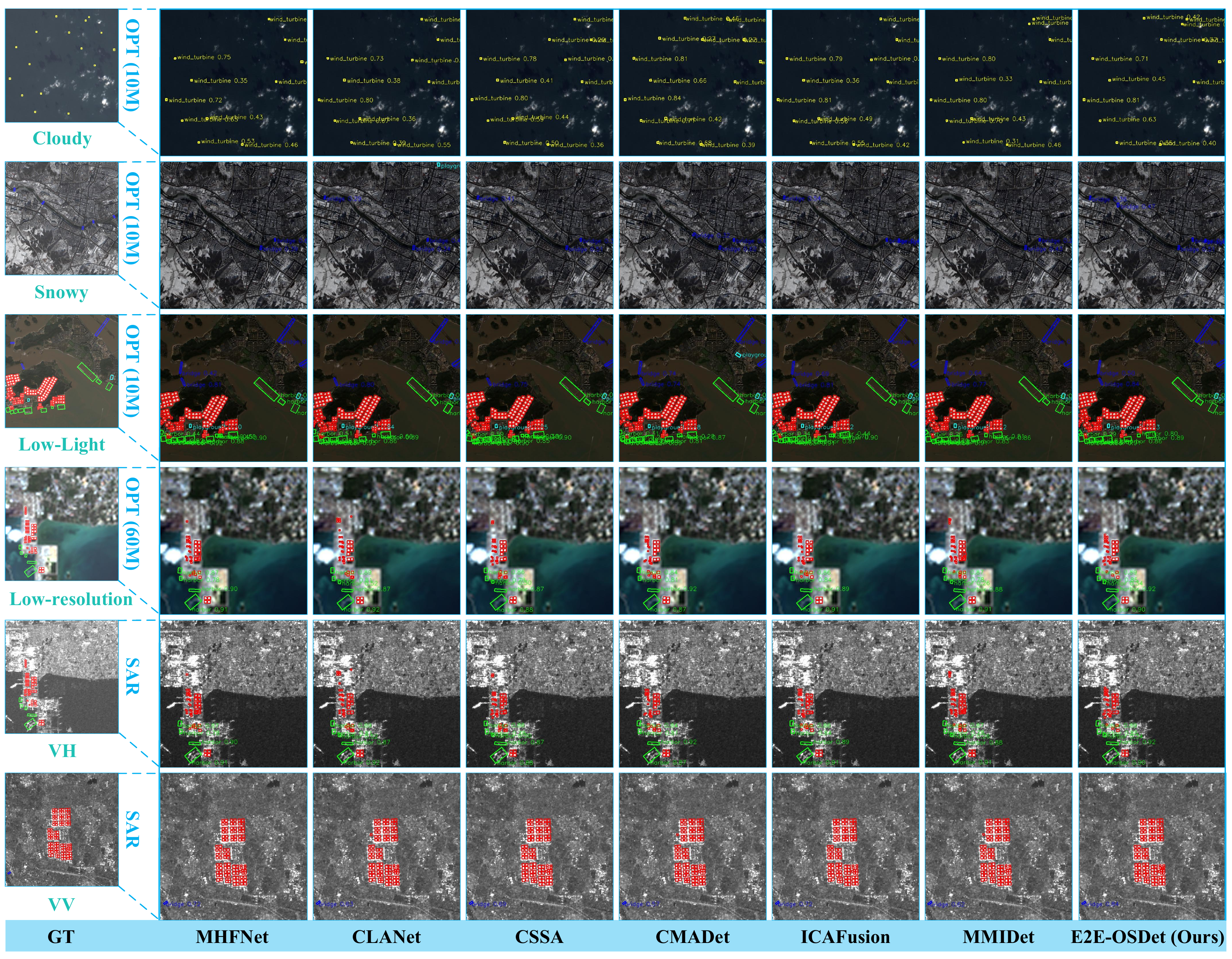}
    \caption {Qualitative comparison of the proposed E2E-OSDet and six fusion methods.}
    \label{vis-detection}
\end{figure*} 

\subsection{Ablation Experiment} 
As shown in Table \ref{albation} (a), we adopt CSSA \cite{cao2023multimodal} as our baseline model, which serves as the foundation for independently evaluating the contribution of each proposed module. The results (see Table \ref{albation} (a), (b), and (c)) indicate that all three modules lead to performance gains, among which the AFM module provides the most substantial improvement in detection accuracy. We further assess the interoperability of the modules. As shown in Table \ref{albation} (e), (f), and (g), the modules are compatible with one another, and their integration does not result in any performance degradation. Finally, we integrate all modules (see Table \ref{albation} (h)), resulting in additional gains in overall model performance.

\textbf{Impact of Filter Augment:} To assess the effectiveness of the FAM, we tested a range of conventional feature descriptors. As shown in Fig. \ref{ablation-experiment} (a) and Fig. \ref{vis-filter-1}, the results indicate that incorporating handcrafted features significantly enhances detection performance. These results mapping pixels into a handcrafted feature space effectively reduces the distribution gap between optical and SAR data. Notably, the Grad feature yields superior performance due to its ability to narrow domain differences and capture multi-scale information. 

\begin{figure*}[!t]	
    \centering
    \includegraphics[width=1.0\linewidth]{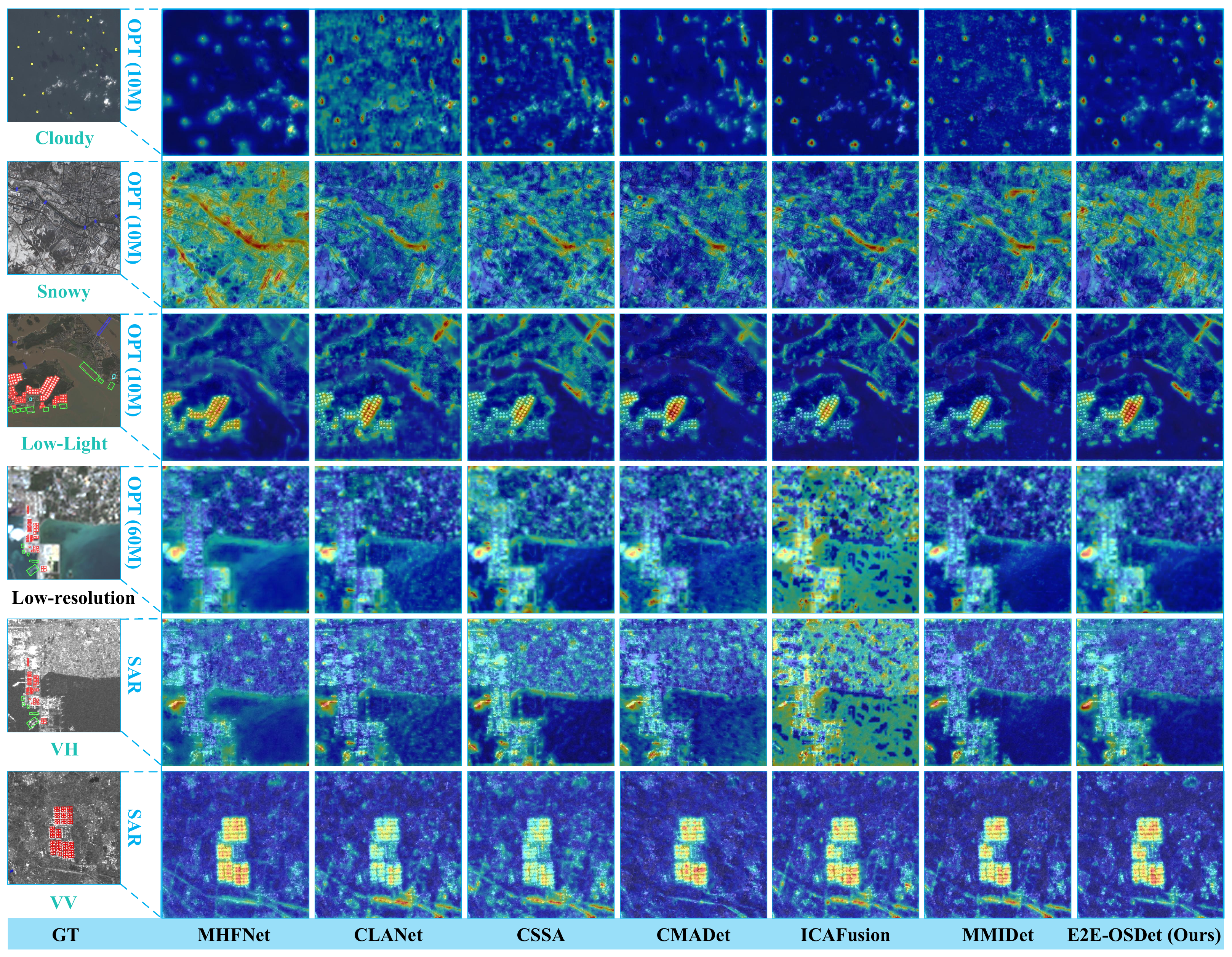}
    \caption {Grad-CAM \cite{selvaraju2017grad} heatmaps of the proposed E2E-OSDet and six fusion methods.
    }
    \label{vis-heatmap}
\end{figure*}

\textbf{Impact of Scanning Mechanisms:} We further investigate the effectiveness of the CMIM module under different scanning mechanisms \cite{he2024mambaad}. As illustrated in Fig. \ref{ablation-experiment} (b)(blue bars), CMIM consistently enhances detection accuracy regardless of the underlying scanning strategy. This observation highlights the robustness and adaptability of CMIM, suggesting that the proposed module can serve as a versatile component that generalizes well across diverse architectural settings rather than being restricted to a particular scanning design.

\textbf{Impact of Feature Proportion and Area Size:} In addition, we investigate the influence of two key design choices: the proportion of handcrafted feature input and the adoption of area attention. As shown in Fig.~\ref{ablation-experiment} (c) and (d), varying these hyperparameters leads to only marginal performance differences, indicating that the overall framework is relatively insensitive to such configurations. This stability suggests that our proposed E2E-OSDet achieves consistent performance without requiring delicate hyperparameter tuning, further underscoring its practicality in real-world deployment.

\subsection{Qualitative Analysis}
% We present visual comparisons of detection results from various multi-source methods. As shown in Fig. \ref{vis-detection}, the proposed E2E-OSDet consistently outperforms other methods in complex scenes. In particular, in regions with limited optical information, such as cloud-covered, snow-covered, low-light, or low-resolution areas, the incorporation of SAR data significantly enhances target localization accuracy. As illustrated in Fig. \ref{vis-heatmap}, our proposed E2E-OSDet produces highly concentrated and spatially aligned attention maps, accurately focusing on target regions while suppressing irrelevant background activations. These experimental results demonstrate the proposed dataset’s effectiveness in evaluating the robustness and adaptability of detection models across diverse scenarios. 
As shown in Fig. \ref{vis-detection} and Fig. \ref{vis-heatmap}, our M4-SAR effectively reveals the strengths and limitations of existing multi-source methods under realistic challenges such as weak cross-modal alignment, scale inconsistency, and angle diversity. Due to coarse geographic alignment and non-simultaneous acquisition, noticeable spatial discrepancies exist between optical and SAR images. Under such conditions, direct fusion methods (e.g., ICAFusion) are prone to mislocalization and orientation errors, while alignment-aware approaches (e.g., CMADet) demonstrate improved robustness. However, even these methods struggle in cases with severe modal shifts or complex backgrounds. In contrast, the E2E-OSDet produces more precise bounding boxes, more accurate angle predictions, and more compact feature activations, indicating stronger cross-modal interaction and background suppression capability. These results validate both the diagnostic value of our dataset in evaluating multi-source fusion strategies and the effectiveness of the E2E-OSDet. More experimental results can be seen in Appendix \ref{Appendix-D}.

\section{Limitations Analysis and Future Work}
Despite the substantial advantages of the proposed M4-SAR in terms of scale, diversity, and scene coverage, certain limitations persist that require further exploration. First, the optical and SAR images are aligned using geographic coordinates rather than pixel-level registration. As a result, precise spatial correspondence between modalities is not guaranteed. This design better reflects realistic cross-sensor acquisition conditions, but it also means that M4-SAR is not suitable for tasks requiring accurate pixel-wise alignment, such as image registration, change detection with strict spatial consistency, or pixel-level fusion. Second, the M4-SAR dataset currently focuses on six representative coastal infrastructure categories. Although these classes exhibit significant variation in scale, aspect ratio, and orientation, the semantic diversity is still limited compared with large-scale natural image benchmarks. Future extensions could incorporate more fine-grained or dynamic categories to further broaden evaluation coverage. These limitations reflect practical trade-offs during dataset construction and highlight potential directions for future improvements.

% \subsection{Future Work}
% Our proposed M4-SAR offers a high-quality benchmark for optical-SAR fusion object detection, with the proposed E2E-OSDet delivering superior performance in this dataset. Future research could investigate the following directions: 1. Broadening the dataset categories to include dynamic and fine-grained targets. 2. Developing lightweight fusion detection architectures for edge deployment compatibility. 3. Implementing temporal alignment and multimodal temporal modeling to address cross-modal acquisition time disparities.
Our proposed M4-SAR establishes a high-quality benchmark for optical–SAR fusion object detection, and the accompanying E2E-OSDet provides a strong baseline for future research. Building upon this foundation, several promising directions merit further investigation. First, expanding the dataset to include more dynamic and fine-grained categories would enhance semantic diversity and broaden its applicability. Second, designing lightweight multimodal fusion architectures could facilitate deployment in resource-constrained or edge computing scenarios. Third, incorporating temporal alignment strategies and multimodal temporal modeling may help mitigate the effects of non-simultaneous acquisitions, further improving cross-modal consistency and robustness.

\section{Conclusion}
\label{sec-7}
This paper presents the M4-SAR dataset and the E2E-OSDet method to tackle two persistent challenges in optical-SAR fusion object detection: the scarcity of high-quality, large-scale datasets and the absence of a unified and effective algorithmic framework. We also develop MSRODet, a comprehensive open-source toolkit for multi-source rotated object detection that integrates mainstream fusion detection algorithms and enables fair performance comparisons on the M4-SAR dataset. We expect that M4-SAR, as a benchmark dataset, will substantially advance research in multi-source remote sensing detection, while E2E-OSDet, as the first fusion framework specifically designed for optical-SAR scenarios, will serve as a strong baseline for future studies.

\section*{Acknowledgements}
This work was supported in part by the National Science Fund of China (No. 62276135, and U24A20330).
% Please insert your acknowledgments here.

% ---- Bibliography ----
%
% BibTeX users should specify bibliography style 'splncs04'.
% References will then be sorted and formatted in the correct style.
%
% \clearpage
\bibliographystyle{splncs04}
\bibliography{main}

\begin{thebibliography}{10}
\providecommand{\url}[1]{\texttt{#1}}
\providecommand{\urlprefix}{URL }
\providecommand{\doi}[1]{https://doi.org/#1}

\bibitem{canny1986computational}
Canny, J.: A computational approach to edge detection. IEEE TPAMI (6),  679--698 (1986)

\bibitem{cao2023multimodal}
Cao, Y., Bin, J., Hamari, J., Blasch, E., Liu, Z.: Multimodal object detection by channel switching and spatial attention. In: CVPR. pp. 403--411 (2023)

\bibitem{cheng2022anchor}
Cheng, G., Wang, J., Li, K., Xie, X., Lang, C., Yao, Y., Han, J.: Anchor-free oriented proposal generator for object detection. IEEE TGRS  \textbf{60},  1--11 (2022)

\bibitem{dalal2005histograms}
Dalal, N., Triggs, B.: Histograms of oriented gradients for human detection. In: CVPR. vol.~1, pp. 886--893 (2005)

\bibitem{ding2021object}
Ding, J., Xue, N., Xia, G.S., Bai, X., Yang, W., Yang, M.Y., Belongie, S., Luo, J., Datcu, M., Pelillo, M., et~al.: Object detection in aerial images: A large-scale benchmark and challenges. IEEE TPAMI  \textbf{44}(11),  7778--7796 (2021)

\bibitem{dong2025fusion}
Dong, W., Zhu, H., Lin, S., Luo, X., Shen, Y., Guo, G., Zhang, B.: Fusion-mamba for cross-modality object detection. IEEE TMM  (2025)

\bibitem{drusch2012sentinel}
Drusch, M., Del~Bello, U., Carlier, S., Colin, O., Fernandez, V., Gascon, F., Hoersch, B., Isola, C., Laberinti, P., Martimort, P., et~al.: Sentinel-2: Esa's optical high-resolution mission for gmes operational services. RSE  \textbf{120},  25--36 (2012)

\bibitem{geng2023focusing}
Geng, P., Lu, X., Hu, C., Liu, H., Lyu, L.: Focusing fine-grained action by self-attention-enhanced graph neural networks with contrastive learning. IEEE TCSVT  \textbf{33}(9),  4754--4768 (2023)

\bibitem{girshick2015fast}
Girshick, R.: Fast r-cnn. In: ICCV. pp. 1440--1448 (2015)

\bibitem{gu2024mamba}
Gu, A., Dao, T.: Mamba: Linear-time sequence modeling with selective state spaces. In: COLM (2024)

\bibitem{he2024mambaad}
He, H., Bai, Y., Zhang, J., He, Q., Chen, H., Gan, Z., Wang, C., Li, X., Tian, G., Xie, L.: Mambaad: Exploring state space models for multi-class unsupervised anomaly detection. NeurIPS  \textbf{37},  71162--71187 (2024)

\bibitem{he2017mask}
He, K., Gkioxari, G., Doll{\'a}r, P., Girshick, R.: Mask r-cnn. In: ICCV. pp. 2961--2969 (2017)

\bibitem{he2023multispectral}
He, X., Tang, C., Zou, X., Zhang, W.: Multispectral object detection via cross-modal conflict-aware learning. In: ACM MM. pp. 1465--1474 (2023)

\bibitem{hu2025exploiting}
Hu, X., Tai, Y., Zhao, X., Zhao, C., Zhang, Z., Li, J., Zhong, B., Yang, J.: Exploiting multimodal spatial-temporal patterns for video object tracking. In: AAAI. vol.~39, pp. 3581--3589 (2025)

\bibitem{Jocher_Ultralytics_YOLO_2023}
Jocher, G., Qiu, J., Chaurasia, A.: {Ultralytics YOLO} (Jan 2023), \url{https://github.com/ultralytics/ultralytics}

\bibitem{li2022yolov6}
Li, C., Li, L., Jiang, H., Weng, K., Geng, Y., Li, L., Ke, Z., Li, Q., Cheng, M., Nie, W., et~al.: Yolov6: A single-stage object detection framework for industrial applications. arXiv preprint arXiv:2209.02976  (2022)

\bibitem{li2024learning}
Li, J., Wang, X., Zhao, H., Zhong, Y.: Learning a cross-modality anomaly detector for remote sensing imagery. IEEE TIP  (2024)

\bibitem{li2024unleashing}
Li, K., Wang, D., Hu, Z., Zhu, W., Li, S., Wang, Q.: Unleashing channel potential: Space-frequency selection convolution for sar object detection. In: CVPR. pp. 17323--17332 (2024)

\bibitem{li2025saratr}
Li, W., Yang, W., Hou, Y., Liu, L., Liu, Y., Li, X.: Saratr-x: Towards building a foundation model for sar target recognition. IEEE TIP  (2025)

\bibitem{li2024predicting}
Li, W., Yang, W., Liu, T., Hou, Y., Li, Y., Liu, Z., Liu, Y., Liu, L.: Predicting gradient is better: Exploring self-supervised learning for sar atr with a joint-embedding predictive architecture. ISPRS  \textbf{218},  326--338 (2024)

\bibitem{li2022oriented}
Li, W., Chen, Y., Hu, K., Zhu, J.: Oriented reppoints for aerial object detection. In: CVPR. pp. 1829--1838 (2022)

\bibitem{li2021deep}
Li, X., Du, Z., Huang, Y., Tan, Z.: A deep translation (gan) based change detection network for optical and sar remote sensing images. ISPRS  \textbf{179},  14--34 (2021)

\bibitem{li2024lsknet}
Li, Y., Li, X., Dai, Y., Hou, Q., Liu, L., Liu, Y., Cheng, M.M., Yang, J.: Lsknet: A foundation lightweight backbone for remote sensing. IJCV pp. 1--22 (2024)

\bibitem{li2024sardet}
Li, Y., Li, X., Li, W., Hou, Q., Liu, L., Cheng, M.M., Yang, J.: Sardet-100k: Towards open-source benchmark and toolkit for large-scale sar object detection. NeurIPS  \textbf{37},  128430--128461 (2024)

\bibitem{lin2023sived}
Lin, X., Zhang, B., Wu, F., Wang, C., Yang, Y., Chen, H.: Sived: A sar image dataset for vehicle detection based on rotatable bounding box. RS  \textbf{15}(11), ~2825 (2023)

\bibitem{lin2025yolo}
Lin, X., Peng, J., Gan, Z., Zhu, J., Liu, J.: Yolo-master: Moe-accelerated with specialized transformers for enhanced real-time detection. arXiv preprint arXiv:2512.23273  (2025)

\bibitem{liu2026cross}
Liu, C., Ma, X., Yang, X., Zhang, Y., Dong, Y.: Como: Cross-mamba interaction and offset-guided fusion for multimodal object detection. Information Fusion  \textbf{125},  103414 (2026)

\bibitem{liu2017high}
Liu, Z., Yuan, L., Weng, L., Yang, Y.: A high resolution optical satellite image dataset for ship recognition and some new baselines. In: ICPRAM. vol.~2, pp. 324--331 (2017)

\bibitem{llerena2021gaussian}
Llerena, J.M., Zeni, L.F., Kristen, L.N., Jung, C.: Gaussian bounding boxes and probabilistic intersection-over-union for object detection. arXiv preprint arXiv:2106.06072  (2021)

\bibitem{lu2025lwganet}
Lu, W., Chen, S.B., Ding, C.H., Tang, J., Luo, B.: Lwganet: A lightweight group attention backbone for remote sensing visual tasks. arXiv preprint arXiv:2501.10040  (2025)

\bibitem{lu2026unravelnet}
Lu, W., Li, H.D., Wang, C., Chen, S.B., Ding, C.H., Tang, J., Luo, B.: Unravelnet: A backbone for enhanced multi-scale and low-quality feature extraction in remote sensing object detection. ISPRS  \textbf{231},  431--442 (2026)

\bibitem{lyu2022rtmdet}
Lyu, C., Zhang, W., Huang, H., Zhou, Y., Wang, Y., Liu, Y., Zhang, S., Chen, K.: Rtmdet: An empirical study of designing real-time object detectors. arXiv preprint arXiv:2212.07784  (2022)

\bibitem{mallat2012group}
Mallat, S.: Group invariant scattering. CPAM  \textbf{65}(10),  1331--1398 (2012)

\bibitem{MMRotate_Contributors_OpenMMLab_rotated_object_2022}
{MMRotate Contributors}: {OpenMMLab rotated object detection toolbox and benchmark} (Feb 2022), \url{https://github.com/open-mmlab/mmrotate}

\bibitem{qingyun2022cross}
Qingyun, F., Zhaokui, W.: Cross-modality attentive feature fusion for object detection in multispectral remote sensing imagery. PR  \textbf{130},  108786 (2022)

\bibitem{selvaraju2017grad}
Selvaraju, R.R., Cogswell, M., Das, A., Vedantam, R., Parikh, D., Batra, D.: Grad-cam: Visual explanations from deep networks via gradient-based localization. In: ICCV. pp. 618--626 (2017)

\bibitem{shen2024icafusion}
Shen, J., Chen, Y., Liu, Y., Zuo, X., Fan, H., Yang, W.: Icafusion: Iterative cross-attention guided feature fusion for multispectral object detection. PR  \textbf{145},  109913 (2024)

\bibitem{song2024misaligned}
Song, K., Xue, X., Wen, H., Ji, Y., Yan, Y., Meng, Q.: Misaligned visible-thermal object detection: a drone-based benchmark and baseline. IEEE TIV  (2024)

\bibitem{WEFT}
Sun, Y., Wang, C., Yang, J., Luo, L.: Small but mighty: Dynamic wavelet expert-guided fine-tuning of large-scale models for optical remote sensing object segmentation. AAAI  (2025)

\bibitem{sun2024united}
Sun, Y., Yang, J., Luo, L.: United domain cognition network for salient object detection in optical remote sensing images. IEEE TGRS  \textbf{62},  1--14 (2024)

\bibitem{tian2025yolov12}
Tian, Y., Ye, Q., Doermann, D.: Yolov12: Attention-centric real-time object detectors. arXiv preprint arXiv:2502.12524  (2025)

\bibitem{tian2019fcos}
Tian, Z., Shen, C., Chen, H., He, T.: Fcos: Fully convolutional one-stage object detection. In: ICCV. pp. 9627--9636 (2019)

\bibitem{torres2012gmes}
Torres, R., Snoeij, P., Geudtner, D., Bibby, D., Davidson, M., Attema, E., Potin, P., Rommen, B., Floury, N., Brown, M., et~al.: Gmes sentinel-1 mission. RSE  \textbf{120},  9--24 (2012)

\bibitem{viola2001rapid}
Viola, P., Jones, M.: Rapid object detection using a boosted cascade of simple features. In: CVPR. vol.~1, pp.~I--I (2001)

\bibitem{wang2024yolov10}
Wang, A., Chen, H., Liu, L., Chen, K., Lin, Z., Han, J., et~al.: Yolov10: Real-time end-to-end object detection. NeurIPS  \textbf{37},  107984--108011 (2024)

\bibitem{wang2025msod}
Wang, C., Fang, W., Li, X., Yang, J., Luo, L.: Msod: A large-scale multiscene dataset and a novel diagonal-geometry loss for sar object detection. IEEE TGRS  \textbf{63},  1--13 (2025)

\bibitem{wang2025cross}
Wang, C., Luo, L., Fang, W., Yang, J.: Cross-modal gaussian localization distillation for optical information guided sar object detection. In: ICASSP. pp.~1--5 (2025)

\bibitem{wang2023category}
Wang, C., Ruan, R., Zhao, Z., Li, C., Tang, J.: Category-oriented localization distillation for sar object detection and a unified benchmark. IEEE TGRS  \textbf{61},  1--14 (2023)

\bibitem{wang2026localized}
Wang, C., Sun, Y., Yang, J., Luo, L.: Localized background-aware generative distillation for enhanced remote sensing object detection. IEEE TCSVT  (2026)

\bibitem{wang2024yolov9}
Wang, C.Y., Yeh, I.H., Mark~Liao, H.Y.: Yolov9: Learning what you want to learn using programmable gradient information. In: ECCV. pp. 1--21 (2024)

\bibitem{wang2022lightweight}
Wang, J., Cui, Z., Jiang, T., Cao, C., Cao, Z.: Lightweight deep neural networks for ship target detection in sar imagery. IEEE TIP  \textbf{32},  565--579 (2022)

\bibitem{wang2019sar}
Wang, Y., Wang, C., Zhang, H., Dong, Y., Wei, S.: A sar dataset of ship detection for deep learning under complex backgrounds. RS  \textbf{11}(7), ~765 (2019)

\bibitem{wei2020hrsid}
Wei, S., Zeng, X., Qu, Q., Wang, M., Su, H., Shi, J.: Hrsid: A high-resolution sar images dataset for ship detection and instance segmentation. IEEE Access  \textbf{8},  120234--120254 (2020)

\bibitem{wu2024fair}
Wu, Y., Suo, Y., Meng, Q., Dai, W., Miao, T., Zhao, W., Yan, Z., Diao, W., Xie, G., Ke, Q., et~al.: Fair-csar: A benchmark dataset for fine-grained object detection and recognition based on single look complex sar images. IEEE TGRS  (2024)

\bibitem{xia2022crtranssar}
Xia, R., Chen, J., Huang, Z., Wan, H., Wu, B., Sun, L., Yao, B., Xiang, H., Xing, M.: Crtranssar: A visual transformer based on contextual joint representation learning for sar ship detection. RS  \textbf{14}(6), ~1488 (2022)

\bibitem{yu2023phase}
Yu, Y., Da, F.: Phase-shifting coder: Predicting accurate orientation in oriented object detection. In: CVPR. pp. 13354--13363 (2023)

\bibitem{yuan2022translation}
Yuan, M., Wang, Y., Wei, X.: Translation, scale and rotation: cross-modal alignment meets rgb-infrared vehicle detection. In: ECCV. pp. 509--525 (2022)

\bibitem{zeng2024mmi}
Zeng, Y., Liang, T., Jin, Y., Li, Y.: Mmi-det: Exploring multi-modal integration for visible and infrared object detection. IEEE TCSVT  (2024)

\bibitem{zhan2025mambasod}
Zhan, Y., Zeng, Z., Liu, H., Tan, X., Tian, Y.: Mambasod: Dual mamba-driven cross-modal fusion network for rgb-d salient object detection. Neurocomputing  \textbf{631},  129718 (2025)

\bibitem{zhang2022domain}
Zhang, C., Feng, Y., Hu, L., Tapete, D., Pan, L., Liang, Z., Cigna, F., Yue, P.: A domain adaptation neural network for change detection with heterogeneous optical and sar remote sensing images. IJAEOG  \textbf{109},  102769 (2022)

\bibitem{zhang2024e2e}
Zhang, J., Cao, M., Yang, X., Xie, W., Lei, J., Li, D., Yang, G., Huang, W., Li, Y.: E2e-mfd: Towards end-to-end synchronous multimodal fusion detection. In: NeurIPS (2024)

\bibitem{zhang2022sefepnet}
Zhang, P., Xu, H., Tian, T., Gao, P., Li, L., Zhao, T., Zhang, N., Tian, J.: Sefepnet: Scale expansion and feature enhancement pyramid network for sar aircraft detection with small sample dataset. IEEE JSTARS  \textbf{15},  3365--3375 (2022)

\bibitem{zhang2020bridging}
Zhang, S., Chi, C., Yao, Y., Lei, Z., Li, S.Z.: Bridging the gap between anchor-based and anchor-free detection via adaptive training sample selection. In: CVPR. pp. 9759--9768 (2020)

\bibitem{zhang2021sar}
Zhang, T., Zhang, X., Li, J., Xu, X., Wang, B., Zhan, X., Xu, Y., Ke, X., Zeng, T., Su, H., et~al.: Sar ship detection dataset (ssdd): Official release and comprehensive data analysis. RS  \textbf{13}(18), ~3690 (2021)

\bibitem{zhang2026mhfnet}
Zhang, W., Zhang, X., Xu, X., Wei, S., Shi, J., Wang, Y., Zeng, T.: Mhfnet: Multimodal hybrid fusion framework for misaligned sar-optical ship detection. ISPRS  \textbf{231},  151--166 (2026)

\bibitem{zhang2025rsar}
Zhang, X., Yang, X., Li, Y., Yang, J., Cheng, M.M., Li, X.: Rsar: Restricted state angle resolver and rotated sar benchmark. In: CVPR. pp. 7416--7426 (2025)

\bibitem{zou2017random}
Zou, Z., Shi, Z.: Random access memories: A new paradigm for target detection in high resolution aerial remote sensing images. IEEE TIP  \textbf{27}(3),  1100--1111 (2017)

\end{thebibliography}

\clearpage
{\Large{\textbf{Supplementary Material: M4-SAR}}}

This appendix is organized as follows:
\begin{itemize}
    \item \textbf{Appendix \ref{Appendix-A}} provides comprehensive details on the structure and annotation protocol of the proposed M4-SAR dataset.
    \item \textbf{Appendix \ref{Appendix-B}} offers additional background for this study, including a review of existing related datasets, remote sensing object detection methods, and multi-source fusion object detection approaches.
    \item \textbf{Appendix \ref{Appendix-C}} presents an in-depth description of the E2E-OSDet architecture, including the Filter Augment Module (FAM), Cross-modal Mamba Interaction Module (CMIM), and Area-attention Fusion Module (AFM), emphasizing their roles in enhancing cross-modal feature fusion.
    \item \textbf{Appendix \ref{Appendix-D}} outlines implementation details, ablation studies, and visualization results, further validating the effectiveness and robustness of the proposed method under diverse conditions.
\end{itemize}

\begin{figure}[!h]	
    \centering
    \includegraphics[width=0.99\linewidth]{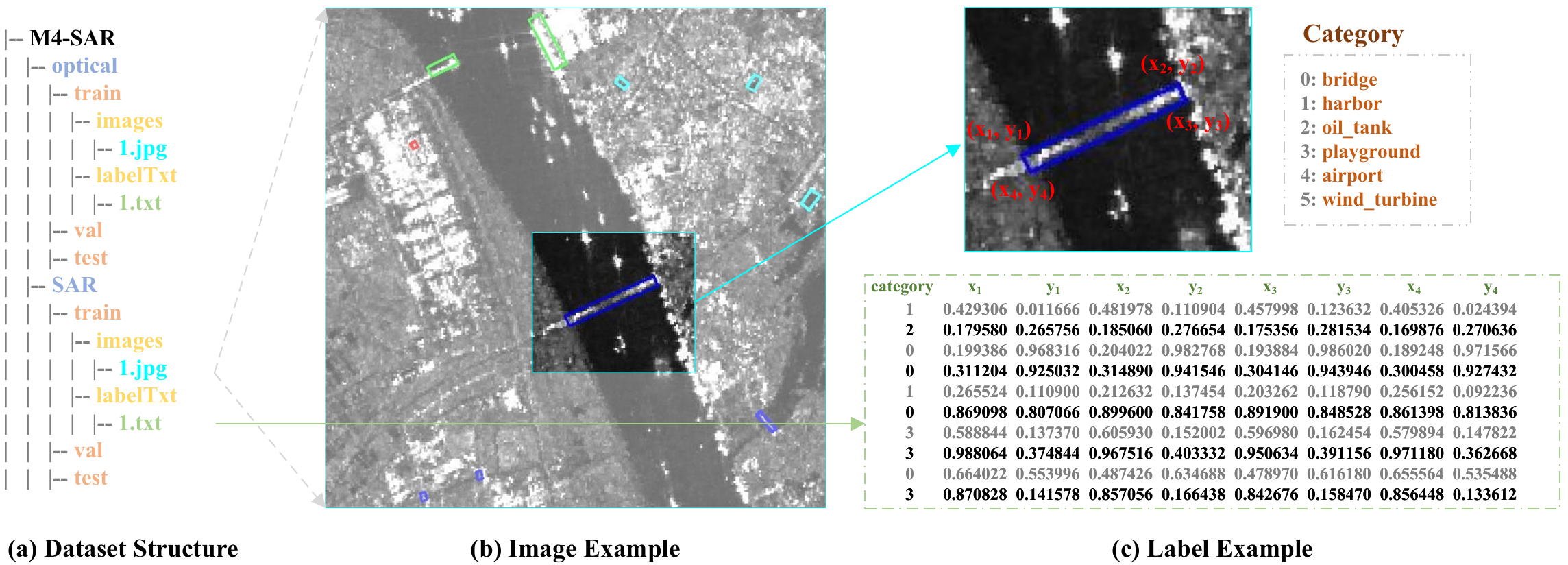}
    \caption {An example of annotation in our M4-SAR dataset.}
    \label{m4-sar-structure}
\end{figure}

\section{More details on the M4-SAR}
\label{Appendix-A}
\subsection{Dataset and Label Format}
\label{Appendix-A1}
We introduce M4-SAR, a novel multi-source remote sensing dataset designed to advance research in multi-source object detection. M4-SAR combines high-resolution optical and synthetic aperture radar (SAR) images, capturing complementary characteristics to address challenges posed by varying illumination, adverse weather conditions, and modality differences. The dataset is carefully organized into three subsets: training, validation, and testing, each comprising paired optical–SAR images with spatially aligned and consistently annotated ground truths.

Annotations are provided in a flexible \texttt{.txt} format, where each object instance is represented by a quadrilateral defined by four vertices: 

$[category, (x_1, y_1), (x_2, y_2), (x_3, y_3), (x_4, y_4)]$ (see Fig. \ref{m4-sar-structure}). 

All coordinates are normalized to the range $[0, 1]$ relative to the image dimensions, allowing precise localization of arbitrarily oriented objects—an essential feature for detecting rotated or skewed targets common in remote sensing images.  

\begin{figure}[!t]	
    \centering
    \includegraphics[width=0.99\linewidth]{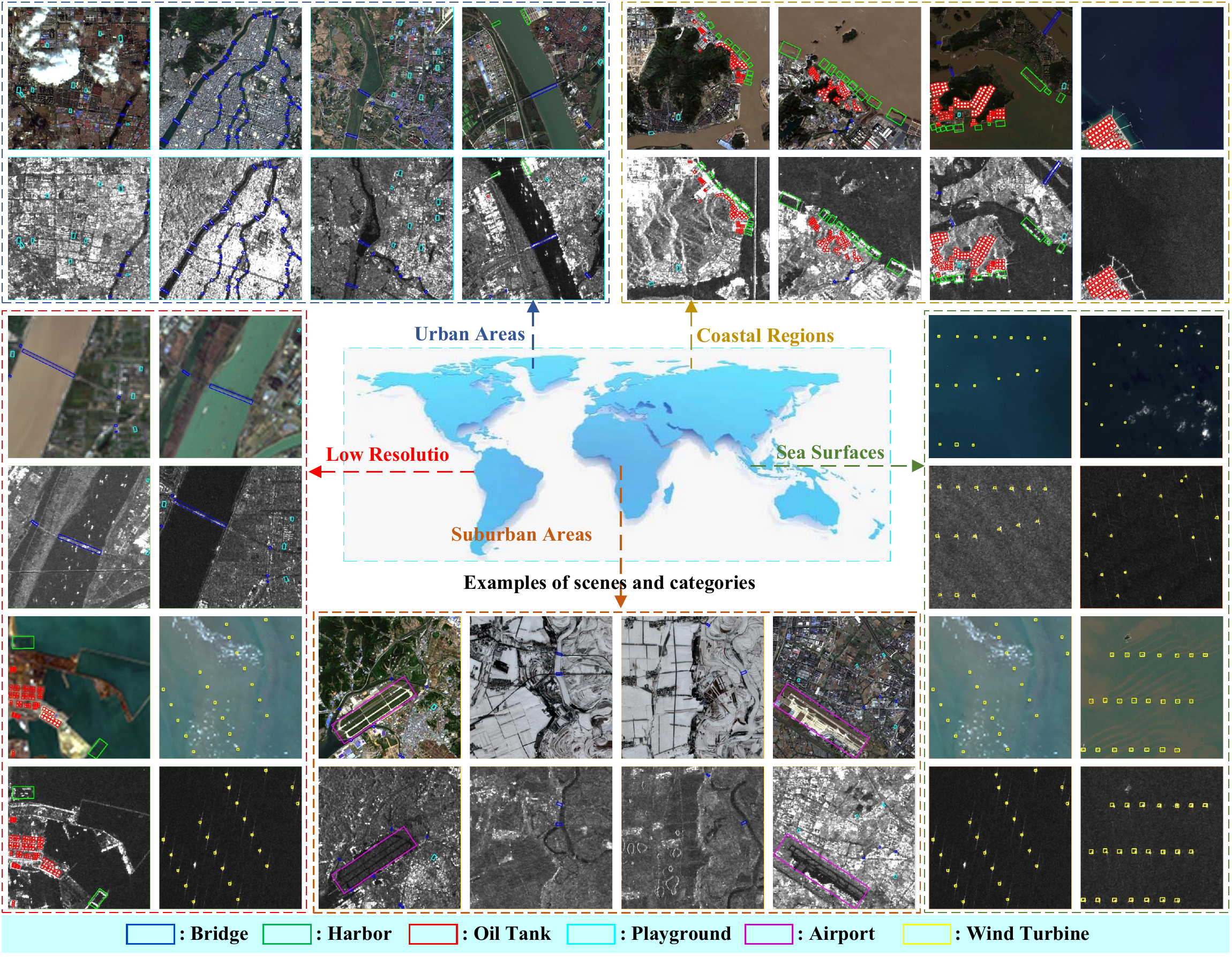}
    \caption {Sample visualization of the M4-SAR dataset.}
    \label{data-vis}
\end{figure}

M4-SAR includes 6 semantically diverse categories: \emph{bridge}, \emph{harbor}, \emph{oil tank}, \emph{playground}, \emph{airport}, and \emph{wind turbine}. These categories cover critical infrastructure and industrial facilities, representing real-world scenes across diverse geographic and environmental conditions (e.g., \emph{playground} has practical significance such as earthquake evacuation and crowd gathering analysis). M4-SAR is designed to drive progress in multi-source object detection by offering a challenging and comprehensive benchmark. Through the integration of optical and SAR modalities with orientation-aware, high-precision annotations, M4-SAR addresses key limitations of existing datasets and provides a solid foundation for developing robust, generalizable models for remote sensing applications. More sample visualization of the M4-SAR dataset is shown in Fig. \ref{data-vis}.

\subsection{Societal Impact and Ethics Consideration}
\label{Appendix-A2}
First, the M4-SAR offers a new perspective on multi-source fusion detection, which can bring about positive societal effects in various applications such as disaster monitoring and national defense. A typical negative societal impact during data collection involves the inclusion of sensitive areas, such as military zones or private properties. We have carefully reviewed this aspect, and we ensured that our acquisition area was limited to non-military and non-private areas.

\section{Related Work}
\label{Appendix-B}
\subsection{Related Datasets}
\label{Appendix-B1}
Progress in SAR object detection is impeded by the high costs of data acquisition and the intricacy of fine-grained annotation. Consequently, existing SAR datasets, including HRSID \cite{wei2020hrsid}, SADD \cite{zhang2022sefepnet}, SSDD \cite{zhang2021sar}, and SIVED \cite{lin2023sived}, are frequently restricted to a limited number of common categories (e.g., ships), generally comprising fewer than 10,000 images with constrained class diversity (Table \ref{dataset}). These constraints impede the development of robust and generalizable models for remote sensing applications. Recent efforts have sought to tackle these challenges. Wang et al. \cite{wang2019sar} introduced the Ship-Dataset, which comprises 39,729 ship patches extracted from Gaofen-3 and Sentinel-1 imagery, extensive in scale but constrained in semantic diversity. Xia et al. \cite{xia2022crtranssar} introduced MSAR dataset, broadening the categories to include airplanes, oil tanks, bridges, and ships. To address data scarcity, Li et al. \cite{li2024sardet} released SARDet-100K, comprising 116,598 images and 245,653 instances across six categories. Wu et al. \cite{wu2024fair} contributed FAIR-CSAR dataset, providing fine-grained multi-domain annotations, while Zhang et al. \cite{zhang2025rsar} enhanced annotation efficiency in RSAR through weakly supervised learning and rotated bounding boxes. In addition, the MAVIC-C \cite{inkawhich20254th} challenge (image classification task) introduced UNICORN V2, a unified optical and radar recognition dataset that includes Earth observation (EO) and synthetic aperture radar (SAR) data. Meanwhile, the MAVIC-T \cite{bowald20253rd} challenge (image translation task) introduced the MAGIC-STACKS dataset to enable the translation of registered aerial images across different sensor modalities.

Despite these advancements, which signify a shift from small-scale, single-category collections to large-scale, multi-class benchmarks, most existing datasets remain restricted to single-source data. Consequently, they cannot exploit complementary multimodal information critical for robust object detection in real-world remote sensing scenarios. Although OGSOD-1.0 \cite{wang2023category} and OGSOD-2.0 \cite{wang2025cross} offer aligned optical-SAR image pairs, their evaluation protocol depends solely on SAR data, limiting their applicability for end-to-end fusion-based detection. Thus, there is an urgent need for a high-quality multi-source dataset that enables comprehensive optical-SAR fusion object detection.

\subsection{Remote Sensing Object Detection}
\label{Appendix-B2}
Remote sensing images, encompassing both optical and synthetic aperture radar (SAR) modalities, pose distinct challenges for object detection. SAR imagery is particularly susceptible to multiplicative speckle noise and structural distortions, which hinder reliable feature extraction. Early approaches sought to address these challenges using handcrafted descriptors, including the Canny edge detector \cite{canny1986computational}, Wavelet Scattering Transform (WST) \cite{mallat2012group}, Haar-like features \cite{viola2001rapid}, Histogram of Oriented Gradients (HOG) \cite{dalal2005histograms}, and Gradient-by-Ratio edge detection (Grad) \cite{li2024predicting}. These methods capture structural information through edge detection, frequency decomposition, or gradient statistics, and have been effectively adapted for both optical and SAR targets. However, their constrained representation capacity and susceptibility to imaging conditions limit their performance in complex remote sensing environments.

With the advent of deep learning, object detection has attained remarkable advancements. Architectures such as R-CNN \cite{girshick2015fast,he2017mask}, YOLO \cite{Jocher_Ultralytics_YOLO_2023,li2022yolov6,wang2024yolov9,wang2024yolov10}, and DETR have exhibited robust generalization across natural images, with numerous variants adapted for remote sensing data. Nevertheless, SAR object detection remains formidable due to speckle noise and the absence of rich semantic cues. To tackle these challenges, recent studies have developed SAR-specific designs. For instance, the SFS-Conv \cite{li2024unleashing} employs fractional Gabor transforms to enhance frequency-domain representations, while SAR-JEPA \cite{li2024predicting} introduces a joint embedding framework that predicts multi-scale gradient features through masked local reconstruction. The large-margin anomaly detection model facilitates cross-modal generalization between optical and SAR data by learning a deviation metric independent of the background distribution. Wang et al. developed a lightweight SAR ship detection network through evolutionary strategies, demonstrating the potential of resource-efficient architectures. Beyond these task-specific advancements, foundational SAR understanding models such as SARATR-X \cite{li2025saratr} reveal that self-supervised pretraining on large-scale SAR datasets produces highly transferable feature representations. 

Despite these advancements, most existing methods are tailored for single-source SAR data and fail to leverage the complementary strengths of auxiliary modalities, such as optical images. This constraint underscores the need for multi-source fusion methods capable of exploiting optical-SAR synergy to reduce noise, enhance semantic context, and improve detection robustness in complex remote sensing scenarios.

\subsection{Multi-source Fusion Object Detection}
\label{Appendix-B3}
Deep neural networks, due to their strong nonlinear modeling capacity, have significantly advanced multi–source fusion object detection \cite{song2024misaligned,cao2023multimodal,shen2024icafusion}. Recent studies have tackled key challenges in cross-modal integration. For example, Yuan et al. \cite{yuan2022translation} and Fang et al. \cite{qingyun2022cross} proposed alignment strategies to mitigate spatial misalignments between heterogeneous modalities. He et al. \cite{he2023multispectral} investigated semantic conflicts and complementarity across sources to enhance fusion effectiveness. Zeng et al. \cite{zeng2024mmi} introduced a contour enhancement module to extract modality-specific spectral features from visible and infrared images, emphasizing spatial diversity and fine-grained structures. Despite these advances, many existing frameworks suffer from complex training procedures, limiting scalability and real-world deployment. To address this, Zhang et al. \cite{zhang2024e2e} proposed E2E-MFD, an end-to-end architecture that streamlines the fusion pipeline through a single-stage training process while achieving strong performance.

However, the majority of prior research has focused on visible–infrared or multispectral-RGB fusion, while optical-SAR fusion remains relatively underexplored. This setting poses unique challenges due to significant disparities in feature distributions and sensing mechanisms. Optical images offer rich color and texture cues but are sensitive to lighting and weather, whereas SAR data provides geometric and structural information resilient to environmental changes. Effectively integrating these complementary properties is essential for robust object detection in complex remote sensing scenarios. The lack of large-scale, high-quality optical-SAR datasets and specialized fusion algorithms highlights an urgent need for new methods tailored to this modality pair. 

\section{More Method Details}
\label{Appendix-C}
\subsection{End-to-End Optical-SAR Object Detection (E2E-OSDet)}
\label{Appendix-C1}
\subsubsection{Overall Framework:}
Multi-source fusion for object detection remains relatively underexplored in the context of optical-SAR integration, and notable challenges persist due to the significant domain discrepancies between the two modalities. As shown in Fig. \ref{overall-network} (b), we propose an end-to-end optical-SAR fusion framework, E2E-OSDet, which explicitly tackles these cross-modal challenges at three complementary levels: data input, domain alignment, and feature fusion.

\subsubsection{Filter Augment Module:}
To address the inherent domain disparities between optical and SAR images, caused by differences in imaging mechanisms and feature distributions, we propose the Filter Augment Module (FAM), a novel preprocessing framework designed to harmonize multi-source inputs for robust object detection. FAM incorporates a suite of classical image filtering operators, including HOG \cite{dalal2005histograms}, Canny \cite{canny1986computational}, Haar \cite{viola2001rapid}, Grad \cite{li2024predicting}, and WST \cite{mallat2012group}. These filters extract complementary low-level cues, such as edges, gradients, and textures, under diverse imaging conditions. By projecting the sparse, low-dimensional features of both optical and SAR images into a high-dimensional, discriminative feature space, FAM reduces cross-modal domain gaps and enhances the alignment of modality-specific characteristics. Specifically, given an optical image $I_O \in \mathbb{R}^{H \times W \times 3}$ and a SAR image $I_S \in \mathbb{R}^{H \times W  \times 3}$, FAM generates a set of enhanced representations $F_{O} \in \mathbb{R}^{H \times W  \times 3}$ and $F_{S} \in \mathbb{R}^{H \times W \times 3}$ for each filter $t \in \{\text{WST}, \text{Canny}, \text{Haar}, \text{HOG}, \text{Grad}\}$, defined as:
\begin{align}
    & F_{O}=I_O + \alpha * \mathbb{FAM}^{t} (I_O),\\
    & F_{S}=I_S + \alpha * \mathbb{FAM}^{t} (I_S),  
     t \in \{\text{WST}, \text{Canny}, \text{Haar}, \text{HOG}, \text{Grad}\},
\end{align}
where $\mathbb{FAM}^{t} (\cdot)$ denotes the $t$-th filter-specific augmentation function, and $\alpha$ represents a set of learnable scalars that balance the contributions of the original and filtered features for the optical and SAR modalities, respectively. This dual-branch enhancement scheme preserves the structural and textural integrity of each modality while injecting noise-robust, multi-scale edge and shape cues derived from classical filters. By aligning feature representations across modalities at the preprocessing stage, FAM facilitates more effective optical-SAR fusion. It addresses the limitations of conventional preprocessing pipelines and significantly enhances the generalization capability of multi-source object detection models under diverse and challenging remote sensing conditions.

\begin{figure*}[t]	
    \centering
    \includegraphics[width=0.99\linewidth]{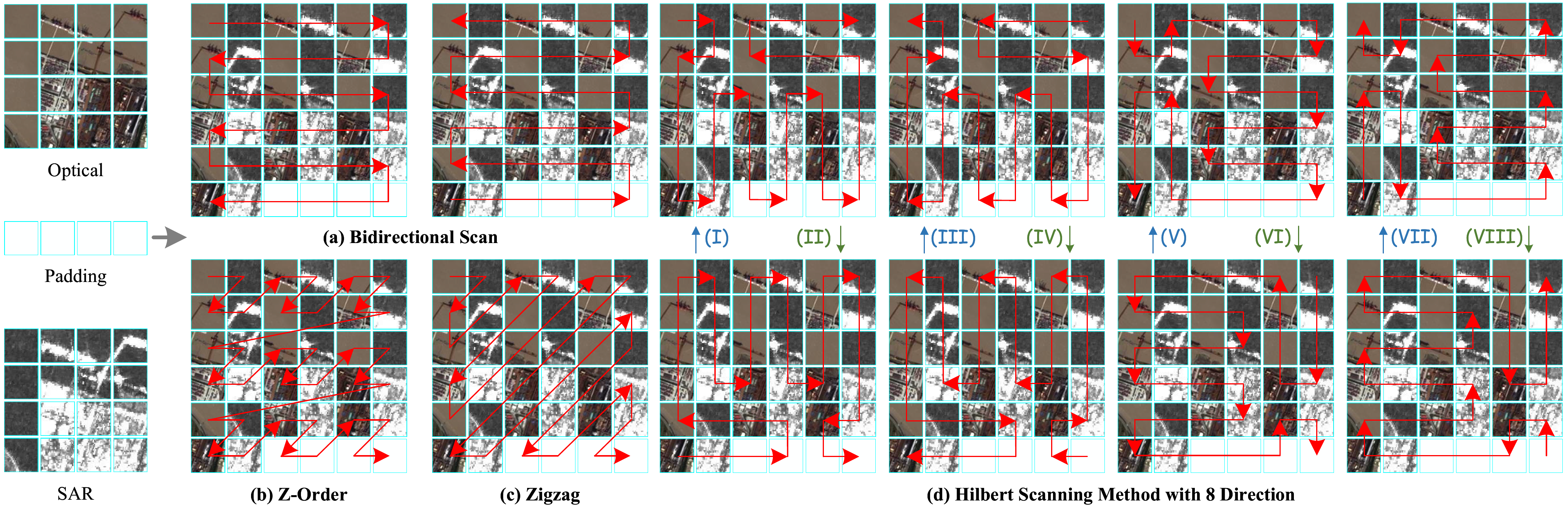}
    \caption {The scanning mechanisms explored in this work include the Bidirectional Scan, Z-Order, Zigzag, and the 8-direction Hilbert Scanning Method.
    }
    \label{scanning-method}
\end{figure*}

\subsubsection{Cross-modal Mamba Interaction Module:} 
Existing multi-source fusion methods primarily emphasize spatial-level feature integration, often neglecting the substantial domain discrepancies between heterogeneous modalities such as optical and SAR images. These domain shifts pose a significant challenge to effective cross-modal fusion. The Mamba framework \cite{gu2024mamba}, with its state-space modeling and efficient sequence processing capabilities, has demonstrated strong potential for capturing long-range dependencies. This makes it well-suited for mitigating inter-modal inconsistencies. However, existing Mamba-based fusion methods \cite{zhan2025mambasod,dong2025fusion} typically involve intricate architectures and computationally intensive interaction designs, which limit their scalability and practical deployment. To overcome these limitations, we propose the Cross-modal Mamba Interaction Module (CMIM), a lightweight yet powerful module tailored to align optical and SAR features by leveraging Mamba’s long-sequence modeling strength. Given paired image features ($F_{Oj}, F_{Sj} \in \mathbb{R}^{\frac{H}{2^{j}}\times \frac{W}{2^{j}} \times C_j}, j \in \{3,4,5\}$), we first flatten them into one-dimensional sequences ($X_{j}\in \mathbb{R}^{1 \times n}$, $Y_{j}\in \mathbb{R}^{1 \times n}$) in row-major order:
\begin{equation}
    X_{j} = [p^o_1, p^o_2, \dots, p^o_{n-1}, p^o_n], Y_{j} = [p^s_1, p^s_2, \dots, p^s_{n-1}, p^s_n], 
\end{equation}
where $p^o_i$ and $p^s_i$ represent patches in optical and SAR imagery, respectively. Traditional methods concatenate these sequences to obtain $Z^{tra}_j \in \mathbb{R}^{1 \times 2n}, j\in \{3,4,5\}$:
\begin{equation}
    Z^{tra}_{j} = [p^o_1, p^o_2, \dots, p^o_{n-1}, p^o_n, p^s_1, p^s_2, \dots, p^s_{n-1}, p^s_n].
\end{equation}
However, these sequence modeling methods often fail to establish explicit local correspondences between modalities, leading to suboptimal cross-modal alignment and feature fusion. 

To address this issue, we introduce an \textbf{Interleaved Input Rearrangement (IIR)} mechanism that enforces fine-grained spatial correspondence by interleaving paired features in a one-to-one fashion. Specifically, given vectorized feature sequences $X_j$ and $Y_j$ from the optical and SAR modalities, respectively, IIR constructs a unified sequence $Z^{IIR}_j \in \mathbb{R}^{1 \times 2n}$ by interleaving the features as:
\begin{equation}
    Z^{IIR}_j = [p^o_1, p^s_1, p^o_2, p^s_2, p^o_3, p^s_3, \dots, p^o_{n-1}, p^s_{n-1}, p^o_n, p^s_n].
\end{equation}

This interleaving ensures that corresponding positional features from both modalities are adjacent, enabling Mamba’s state-space modeling to directly capture semantic and structural alignments, thus enhancing domain coherence. The resulting interleaved sequence is subsequently fed into a directional scanning mechanism (see Fig. \ref{scanning-method} \cite{hu2025exploiting}), which applies Mamba blocks along both horizontal and vertical directions. This bi-directional sequence modeling allows CMIM to capture long-range dependencies while preserving fine-grained local correspondence, facilitating deeper cross-modal interaction. Finally, the output sequence is reshaped back into the 2D spatial domain and fused with residual skip connections to retain original modality-specific information. Through this design, CMIM effectively bridges modality gaps and enhances the robustness of multi-source object detection under challenging remote sensing conditions.
\begin{align}
   & Z^{scan}_j = \mathbb{SCAN}^m(Z^{IIR}_j), m \in\{\text{Bid, Z-Order, Zigzag, Hilbert}\}.
\end{align}
The CMIM then transforms the scanned sequence into enhanced feature representations:
\begin{equation}
    \{F'_{Oj}, F'_{Sj}\} = \mathbb{CMIM} (Z^{scan}_j), \quad j \in\{3, 4, 5 \}.
\end{equation}
By positioning optical and SAR features adjacent along the temporal axis, the Interleaved Input Rearrangement (IIR) mechanism enables synchronous semantic enhancement within short time steps, effectively mitigating the risk of feature divergence during training. This temporal alignment facilitates efficient parameter sharing across modalities, stabilizes the training process, and promotes consistent gradient flow. As a result, IIR significantly enhances cross-modal feature alignment and consistently outperforms conventional fusion strategies in complex remote sensing scenarios.

\subsubsection{Area-attention Fusion Module:}
To enhance the discriminative capacity of fused optical-SAR features across spatially diverse regions, we propose the Area-attention Fusion Module (AFM), a novel mechanism that emphasizes salient local structures and amplifies responses in critical target regions. AFM exploits the complementary characteristics of optical and SAR modalities, enabling robust and adaptive feature integration for multi-source object detection. Inspired by the efficient spatial modeling techniques in \cite{tian2025yolov12,geng2023focusing}, AFM partitions the input feature map of resolution $(H, W)$ into $k$ non-overlapping blocks of size $(H/k, W)$ or $(H, W/k)$, without requiring explicit window segmentation. This lightweight partitioning is implemented via a simple reshaping operation, preserving spatial continuity while substantially improving computational efficiency. The fused feature representation is computed as:
\begin{equation}
    F^{fus}_{j} = \mathbb{AFM} (F'_{Oj}+F_{Oj}, F'_{Sj}+F_{Sj}), \quad j \in\{3, 4, 5 \},
\end{equation}
where $F'_{Oj}$ and $F'_{Sj}$ denote the optical and SAR features. By adaptively weighting localized features, AFM improves the model’s capability to detect targets in complex remote sensing scenes characterized by varying object scales and cluttered backgrounds. As illustrated in Fig. \ref{overall-network} (c), AFM achieves efficient and effective cross-modal fusion, surpassing conventional spatial fusion strategies that often overlook regional saliency and incur higher computational overhead.

\subsection{Loss Function}
\label{Appendix-C2}
In the proposed E2E-OSDet framework, we employ a composite loss function comprising Binary Cross-Entropy (BCE) and Probabilistic IoU (ProbIoU) \cite{llerena2021gaussian} to jointly optimize classification and localization. The overall loss is defined as:
\begin{equation}
    \mathcal{L}_{total} = \mathcal{L}_{reg} + \mathcal{L}_{dfl} + \mathcal{L}_{cls},
\end{equation}
where $\mathcal{L}_{reg}$ denotes the bounding box regression loss, $\mathcal{L}_{dfl}$ represents the distribution focus loss, and $\mathcal{L}_{cls}$ corresponds to the classification loss. This formulation balances localization accuracy and classification confidence, ensuring end-to-end optimization of the optical-SAR fusion detection pipeline.

\section{More Experimental Results}
\label{Appendix-D}
\subsection{More Implementaion Details}
\label{Appendix-D1}
To facilitate robust evaluation and accelerate progress in multi-source object detection, we introduce \textbf{MSRODet}, a comprehensive and extensible toolbox built upon the Ultralytics framework~\cite{Jocher_Ultralytics_YOLO_2023}. MSRODet integrates a wide range of fusion-based detectors, including MHFNet \cite{zhang2026mhfnet}, CFT~\cite{qingyun2022cross}, CLANet~\cite{he2023multispectral}, CSSA~\cite{cao2023multimodal}, CMADet~\cite{song2024misaligned}, ICAFusion~\cite{shen2024icafusion}, MMIDet~\cite{zeng2024mmi}, and our proposed E2E-OSDet, alongside single-source methods such as YOLOv5~\cite{Jocher_Ultralytics_YOLO_2023}, YOLOv6~\cite{li2022yolov6}, YOLOv8~\cite{Jocher_Ultralytics_YOLO_2023}, YOLOv9~\cite{wang2024yolov9}, YOLOv10~\cite{wang2024yolov10}, YOLOv11~\cite{Jocher_Ultralytics_YOLO_2023}, YOLOv12~\cite{tian2025yolov12}, YOLO26~\cite{Jocher_Ultralytics_YOLO_2023} and YOLO-Master~\cite{lin2025yolo}. To ensure fair comparisons, all multi-source fusion methods adopt YOLOv11-S as the unified backbone and neck architecture. We also extend these methods to other YOLO variants to evaluate their framework compatibility while maintaining consistent model capacity.

For oriented object detection, crucial in remote sensing applications, we replace the standard detection heads with the oriented bounding box (OBB) head from YOLOv8~\cite{Jocher_Ultralytics_YOLO_2023}, enabling accurate localization of arbitrarily rotated targets. Training is performed using a momentum-based SGD optimizer (initial learning rate: 0.01, momentum: 0.937, weight decay: 5e-4) for 300 epochs with a batch size of 64. All input images are resized to 512$\times$512, and experiments are conducted on two NVIDIA RTX 4090 GPUs with mixed-precision (FP16) training. To eliminate pretraining bias, all fusion models are trained from scratch.

For other methods, we also evaluate single-source baselines implemented via the MMRotate toolkit~\cite{MMRotate_Contributors_OpenMMLab_rotated_object_2022}, including Rotated FCOS~\cite{tian2019fcos}, Rotated ATSS~\cite{zhang2020bridging}, Oriented RepPoint~\cite{li2022oriented}, RTMDet~\cite{lyu2022rtmdet}, PSC~\cite{yu2023phase}, and LSKNet~\cite{li2024lsknet}. All methods use a ResNet-50 backbone pre-trained on ImageNet, and are optimized using AdamW (learning rate: 2e-4, weight decay: 0.05) for 72 epochs with a batch size of 32 and the same 512$\times$512 input resolution. This unified and standardized setup ensures fair, reproducible, and extensible comparisons across a wide spectrum of single-source and multi-source object detection methods, establishing MSRODet as a robust benchmark platform for optical–SAR fusion in remote sensing.

\begin{table}[h]
\centering
\caption{Scores from human expert (3,000 samples).}
\fontsize{8.0pt}{10pt}\selectfont
\setlength{\tabcolsep}{1.0mm}
\begin{tabular}{c|cccc|cccc}
\hline
\multicolumn{1}{c}{\multirow{2}{*}{\textbf{Scores}}} & \multicolumn{4}{|c|}{\textbf{RS Experts}} & \multicolumn{4}{c}{\textbf{Non-RS Experts}} \\ \cline{2-9} 
\multicolumn{1}{c|}{}  & \textbf{1}    & \textbf{2}    & \textbf{3}    & \textbf{4}    & \textbf{1}    & \textbf{2}    & \textbf{3}    & \textbf{4}    \\ \hline
All         & 98.5\% & 98.1\% & 97.4\% & 99.3\% & 99.3\% & 99.9\% & 99.5\% & 98.4\% \\ \hline
Average & \multicolumn{4}{c|}{98.325\%}  & \multicolumn{4}{c}{99.275\%}  \\ \hline
\end{tabular}
\label{human expert}
\end{table}

\begin{table}[h]
\centering
\caption{Results using mixed optical and SAR data.}
\fontsize{8.0pt}{10pt}\selectfont
\setlength{\tabcolsep}{1.0mm}
\begin{tabular}{c|cc|cc|cc}
\hline
\multirow{2}{*}{\textbf{Method}} & \multicolumn{2}{c|}{\textbf{Optical}} & \multicolumn{2}{c|}{\textbf{SAR}} & \multicolumn{2}{c}{\textbf{Optical + SAR}} \\ \cline{2-7} 
        & \textbf{AP}$_{50}$ $\uparrow$ & \textbf{mAP} $\uparrow$  & \textbf{AP}$_{50}$ $\uparrow$ & \textbf{mAP} $\uparrow$  & \textbf{AP}$_{50}$ $\uparrow$ & \textbf{mAP} $\uparrow$  \\ \hline
YOLOv11 & 79.2\% & 54.3\% & 71.3\% & 44.7\% & 72.1\%    & 47.2\%    \\
YOLO26  & 78.4\% & 55.4\% & 73.3\% & 48.2\% & 71.1\%    & 47.3\%    \\ 
YOLO-Master  & 80.0\% & 56.3\% & 73.9\% & 48.5\% & 73.2\%    & 49.0\%    \\
\hline \hline
E2E-OSDet(Ours)     & -      & -      & -      & -      & \textbf{85.7}\% & \textbf{61.4}\% \\ \hline
\end{tabular}
\label{mixeddata}
\end{table}

\subsection{Human Expert Score of M4-SAR}
We randomly sampled 3,000 samples from the dataset and invited 4 Remote Sensing (RS) experts and 4 Non-Remote Sensing (Non-RS) experts to conduct a subjective evaluation of the annotation quality (correct annotations were counted as correct, while both missing and incorrect annotations were counted as errors). As shown in Table \ref{human expert}, the average scores exceed 98\% in all cases.

\subsection{Mixed Optical-SAR Data for Single-Detector Training}
As shown in Table \ref{mixeddata}, directly mixing optical-SAR data for training leads to performance degradation due to the significant domain gap between the two modalities. For example, YOLO26 achieves only 47.3\% mAP on the mixed data, which is lower than its performance on optical (55.4\% mAP) and SAR (48.2\% mAP) data. In contrast, our proposed E2E-OSDet achieves a much higher mAP of 61.4\%, demonstrating that a specially designed fusion architecture is crucial for fully exploiting the complementary characteristics of optical and SAR data. 

\subsection{ M4-SAR Dataset Split Strategy}
M4-SAR is partitioned using a random split strategy with a training/validation/test ratio of 5:2:3.

\subsection{Visualization Results for Handcrafted Feature}
\label{Appendix-D2}
\begin{figure}[!t]	
    \centering
    \includegraphics[width=1.0\linewidth]{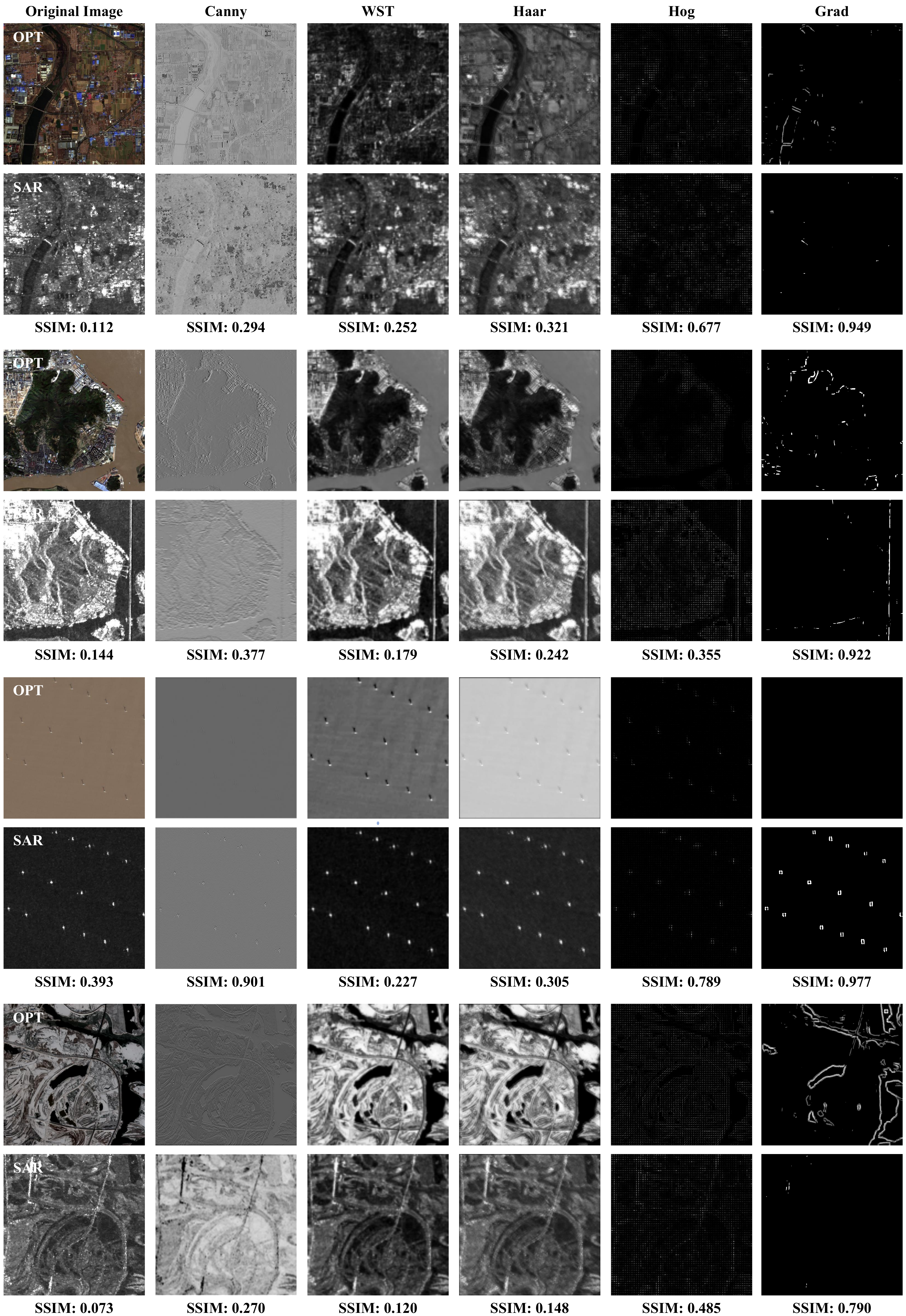}
    \caption {Visualization of handcrafted features on SAR images, with SSIM denoting structural similarity. Features are average-pooled and represented as a single channel for visualization.}
    \label{vis-filter-2}
\end{figure}

To assess the effectiveness of handcrafted features in mitigating domain disparities between optical and SAR images, we examine a set of classical feature descriptors, including the Canny Edge Detector~\cite{canny1986computational}, Wavelet Scattering Transform (WST)~\cite{mallat2012group}, Haar-like Features~\cite{viola2001rapid}, Histogram of Oriented Gradients (HOG)~\cite{dalal2005histograms}, and Gradient by Ratio Edge (Grad)~\cite{li2024predicting}. We employ the Structural Similarity Index Measure (SSIM) to quantitatively evaluate the structural alignment between optical and SAR images in the handcrafted feature space. As shown in Fig.~\ref{vis-filter-2}, these handcrafted descriptors significantly enhance cross-modal structural similarity compared to the original pixel domain, effectively bridging the intrinsic imaging differences between modalities.

Notably, the Grad feature space~\cite{li2024predicting} achieves the highest SSIM scores, significantly outperforming other descriptors by capturing noise-resilient edge information that is particularly critical for SAR images. This result highlights Grad’s superior capacity for aligning multi-source features, especially under challenging conditions characterized by multiplicative speckle noise in SAR data. Motivated by this observation, our method prioritizes the integration of Grad-based handcrafted features within the Filter Augment Module (FAM) to enhance the structural alignment and fusion of optical–SAR representations, thereby improving the robustness and discriminability of multi-source object detection in complex remote sensing environments.

\subsection{Detection Result Visualizations}
\label{Appendix-D3}
\begin{figure}[!t]	
    \centering
    \includegraphics[width=1.0\linewidth]{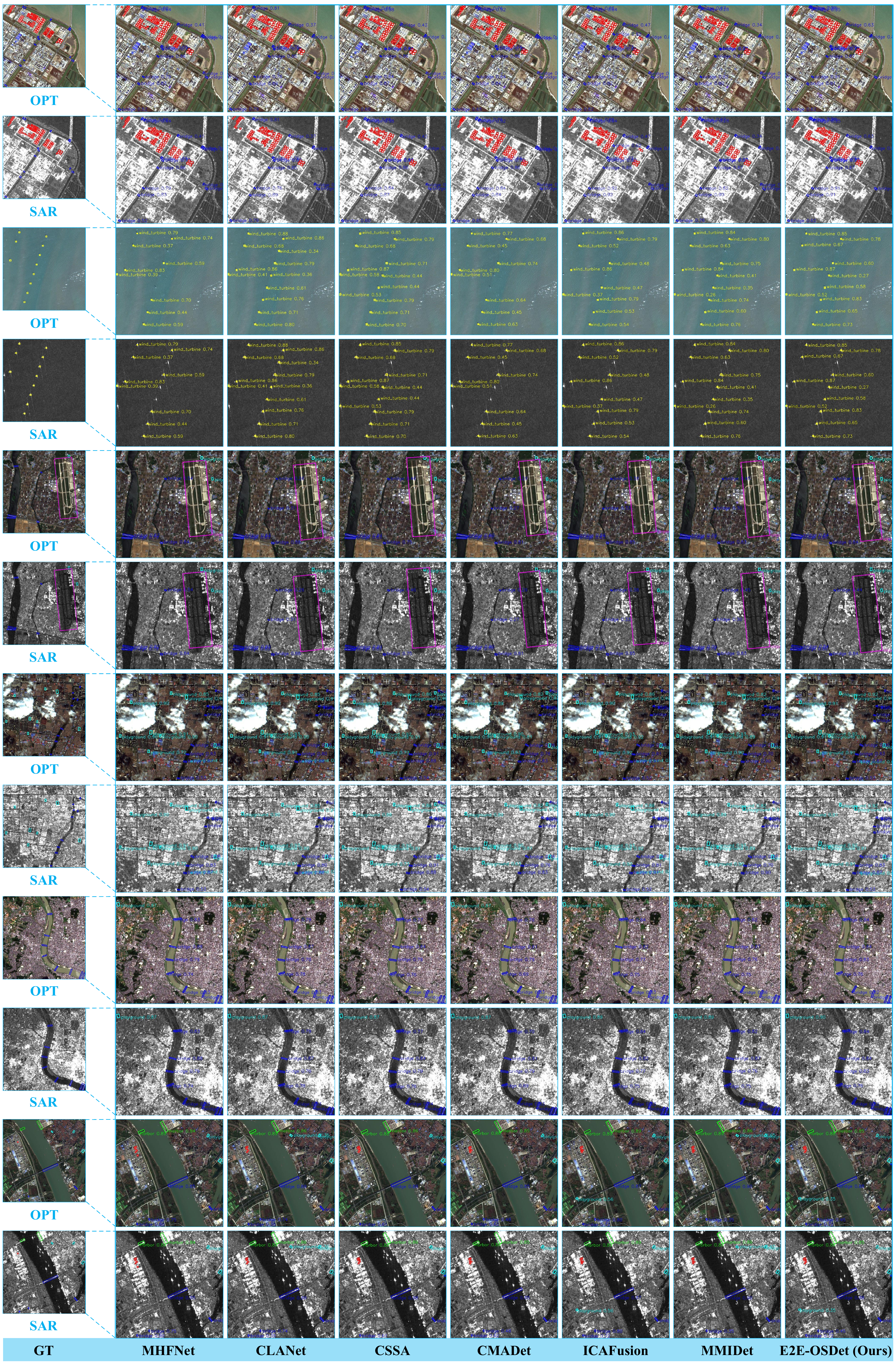}
    \caption {Comparison of detection results across different methods (zoom in for detail).}
    \label{vis-detection-2}
\end{figure}

To elucidate the effectiveness of our multi-source fusion framework for optical–SAR object detection, we present a comprehensive visual comparison of detection outcomes across diverse input modalities and fusion strategies. As shown in Fig.~\ref{vis-detection-2}, our method effectively harnesses the complementary characteristics of optical (rich texture and color) and SAR (robust structural information) images, yielding sharper target boundaries and higher detection confidence relative to existing fusion baselines. In particularly challenging scenarios—including low-contrast environments, cluttered urban scenes, and adverse weather conditions—our approach exhibits strong resilience, significantly improving both recognition accuracy and localization precision. These visual comparisons highlight the critical role of cross-modal alignment and deep semantic integration, underscoring the practical applicability and generalization capability of our method in real-world remote sensing tasks.

\subsection{Heatmap Visualizations}
\label{Appendix-D4}
\begin{figure}[!t]	
    \centering
    \includegraphics[width=1.0\linewidth]{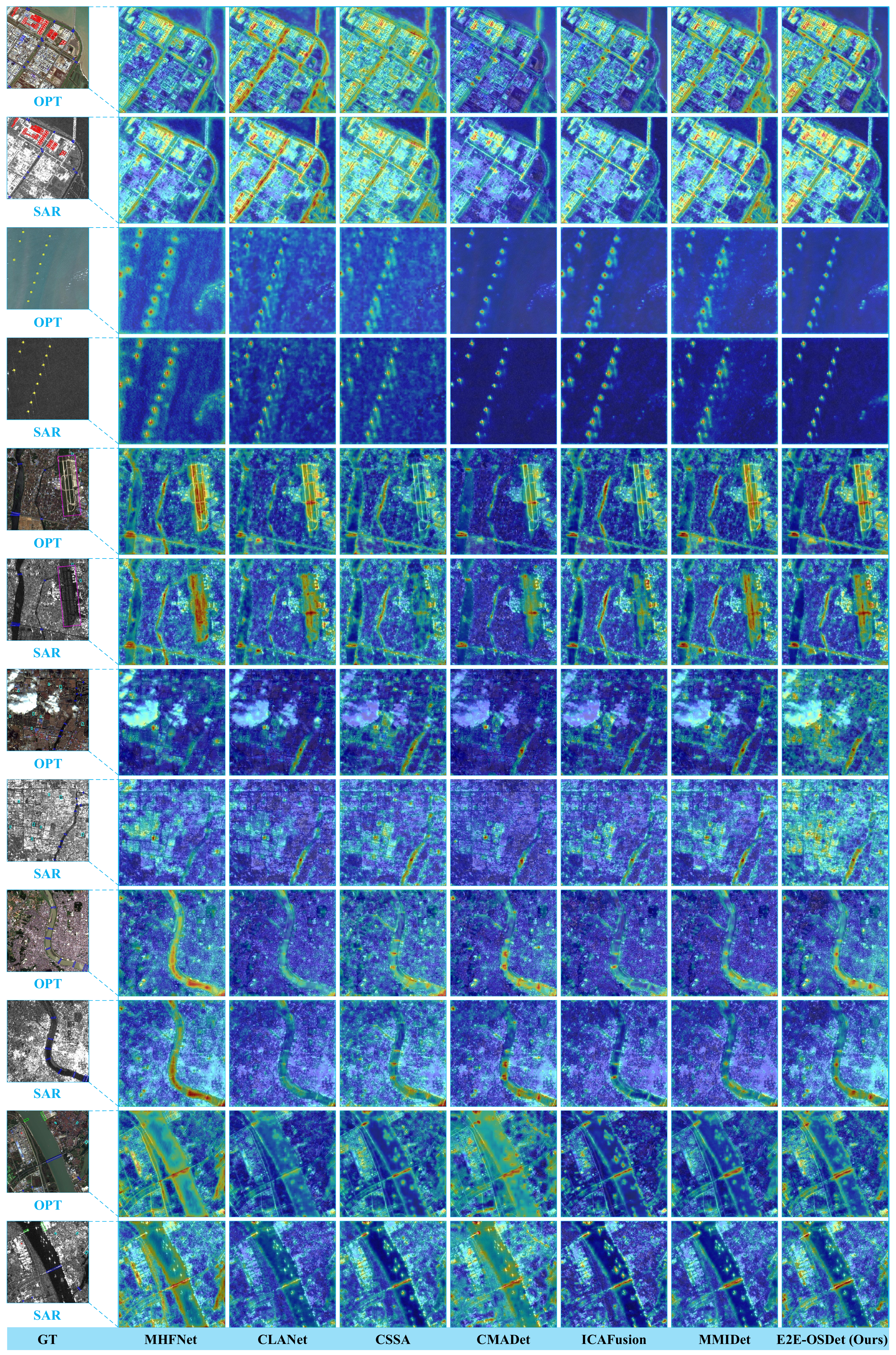}
    \caption {Grad-CAM \cite{selvaraju2017grad} heatmaps of the proposed E2E-OSDet and six fusion methods.
    }
    \label{vis-heatmap-1}
\end{figure}

To assess the effectiveness of our model in leveraging multi-source information, we present attention heatmap visualizations comparing the response distributions of various methods on representative samples from our optical–SAR dataset. As illustrated in Fig.~\ref{vis-heatmap-1}, our proposed E2E-OSDet produces highly concentrated and spatially aligned attention maps, accurately focusing on target regions while suppressing irrelevant background activations. This indicates a strong ability to extract semantically meaningful and spatially coherent features from heterogeneous modalities. In contrast, baseline fusion methods yield diffuse or misaligned attention responses, often exhibiting attention drift or overlooking low-contrast objects, especially in cluttered or noisy scenes. These qualitative results emphasize the importance of cross-modal alignment and semantic integration, and demonstrate that E2E-OSDet significantly improves attention precision, detection robustness, and generalization to complex remote sensing environments.

\begin{table}[]
\centering
\caption{Quantitative results on the OGSOD-1.0 and OGSOD-2.0.}
\fontsize{8.0pt}{10pt}\selectfont
\setlength{\tabcolsep}{2.3mm}
\begin{tabular}{l|cc|ccc|ccc}
\Xhline{1.0pt} 
\multirow{2}{*}{\textbf{Method}} & \multirow{2}{*}{\textbf{O}} & \multirow{2}{*}{\textbf{S}} & \multicolumn{3}{c|}{\textbf{OGSOD-1.0\cite{wang2023category}}} & \multicolumn{3}{c}{\textbf{OGSOD-2.0\cite{wang2025cross}}} \\ \cline{4-9} 
                       && & $\textbf{AP}_{50}$  & $\textbf{AP}_{75}$  & $\textbf{mAP}$     & $\textbf{AP}_{50}$  & $\textbf{AP}_{75}$  & $\textbf{mAP}$     \\ \hline \hline
YOLOv11 \cite{Jocher_Ultralytics_YOLO_2023}                  &$\checkmark$& {\color{black!30}\ding{55}} & 90.8     & 66.0     & 61.5    & 88.2     & 62.6     & 58.5    \\
YOLOv11 \cite{Jocher_Ultralytics_YOLO_2023}                  &{\color{black!30}\ding{55}}&$\checkmark$  & 76.0     & 48.0     & 47.8    & 73.8     & 46.2     & 45.9    \\
\hline
CFT \cite{qingyun2022cross}                   &$\checkmark$ &   $\checkmark$  & 92.9     & 72.1     & 65.3    & 90.8     & 67.1     & 61.4    \\
CLANet \cite{he2023multispectral}              &$\checkmark$ &   $\checkmark$    & 93.5     & 72.9     & 66.3    & 91.6     & 68.7     & 63.0    \\
CSSA \cite{cao2023multimodal}                  &$\checkmark$ &   $\checkmark$  & 92.1     & 69.3     & 63.3    & 89.7     & 64.2     & 59.8    \\
CMADet \cite{song2024misaligned}               &$\checkmark$ &   $\checkmark$  & 91.3     & 67.6     & 62.9    & 88.1     & 60.9     & 57.7    \\
ICAFusion \cite{shen2024icafusion}             &$\checkmark$ &   $\checkmark$  & 93.5     & 72.7     & 66.1    & 90.7     & 68.2     & 62.2    \\
MMIDet \cite{zeng2024mmi}                &$\checkmark$ &   $\checkmark$  & 93.5     & 73.5     & 66.7    & 91.6     & 68.6     & 63.0    \\ \hline \hline
\rowcolor{black!5}
\textbf{E2E-OSDet}             &$\checkmark$ &   $\checkmark$  &  \textbf{93.6}     &  \textbf{74.3}     &  \textbf{67.3}    &  \textbf{92.3}     &  \textbf{71.8}     &  \textbf{65.1}    \\ \Xhline{1.0pt} 
\end{tabular}
\label{ogosd-result}
\end{table}

\subsection{Expanded Application of E2E-OSDet}To further assess the generalization capability of the E2E-OSDet framework, we conducted additional experiments on the OGSOD-1.0 \cite{wang2023category} and OGSOD-2.0 \cite{wang2025cross} datasets (300 epochs, input resolution of 256$\times$256). As shown in Table~\ref{ogosd-result}, E2E-OSDet attains the highest performance on OGSOD-1.0, achieving an $mAP$ of 67.3\% and surpassing representative multimodal methods, including CLANet, ICAFusion, and MMIDet. These results demonstrate the robustness and generalizability of E2E-OSDet across various benchmarks.

\end{document}